%% file: arxiv_v2.tex
\documentclass[10pt]{article} 
\usepackage[preprint]{tmlr}

\usepackage{hyphenat}
\PassOptionsToPackage{numbers, compress}{natbib}
\input{math_commands.tex}

\usepackage[utf8]{inputenc}
\usepackage[T1]{fontenc}
\usepackage{hyperref}
\usepackage{url}
\usepackage{booktabs}
\usepackage{amsfonts}
\usepackage{amsmath,amssymb,amsthm,mathtools}
\usepackage{nicefrac}
\usepackage{microtype}
\usepackage{xcolor}
\newcommand{\rebuttal}[1]{#1}
\usepackage{enumitem}
\usepackage{algorithm}
\usepackage{algpseudocode}
\usepackage{subcaption}
\usepackage[normalem]{ulem}
\title{Structure-Preserving Correction Learning for Sparse Bayesian Inference in Brain Source Imaging}
\author{
Marco Morik$^{1,2,\dagger}$
Xiao Ruiting$^{2}$
Shinichi Nakajima$^{1,2,3}$
Stefan Haufe$^{2,4,5}$
Ismail Huseynov$^{2,4,\ddagger}$ \hspace{-4em}
\\[1em]
\vspace{-1em}
\\ 
$^{1}$Berlin Institute for the Foundations of Learning and Data (BIFOLD), Germany \\
$^{2}$Technische Universität Berlin, Germany \\
$^{3}$RIKEN Center for Advanced Intelligence Project (AIP), Japan \\
$^{4}$Physikalisch-Technische Bundesanstalt, Germany \\
$^{5}$Charité -- Universitätsmedizin Berlin, Germany \\
$^\dagger$ \texttt{m.morik@tu-berlin.de}, $^\ddagger$ \texttt{ismail.huseynov@ptb.de}
}

\begin{document}
\raggedbottom

\maketitle

\begin{abstract}
Classical sparse Type-II Bayesian methods for M/EEG brain imaging support joint estimation of source and noise hyperparameters, but rely on fixed iterative update rules. Although these updates are principled and interpretable, their dynamics are not data-driven. We propose to learn the update mechanism itself while preserving the underlying Bayesian structure by unfolding a classical joint hyperparameter-learning solver into a trainable neural network architecture whose layers mirror the original iterations. The resulting framework is initialized to recover the classical solver exactly before training and is enriched through progressively more expressive \textit{residual correction learning} mechanisms, ranging from learnable biases to adaptive multilayer perceptron and attention-based contextual refinements. In this way, training does not replace Bayesian inference with a black-box predictor, but instead learns structured correction terms while retaining the interpretability and model-based character of the original update dynamics. Structured correction learning therefore aims to improve empirical reconstruction performance without replacing the original model-based inference mechanism. Experimental results show that the learned correction variants improve reconstruction performance and convergence behavior over the baseline unfolded solver while preserving its algorithmic transparency.
\end{abstract}

\section{Introduction}
\label{sec:introduction}
Electroencephalography (EEG) and magnetoencephalography (MEG) provide noninvasive measurements of brain activity with millisecond temporal resolution, \rebuttal{that can be used to localize epileptogenic activity, interpreting functional generators, or studying individual brain dynamics \cite{megevand2020electric}.} However, reconstructing the underlying neural sources from these sensor recordings remains a severely ill-posed inverse problem. In practical M/EEG source imaging, the number of candidate sources is typically much larger than the number of sensors, the lead-field operator is spatially smoothing, and different source configurations can produce similar measurements. Consequently, reconstruction quality depends strongly on the prior model, the noise model, and the inference algorithm used to estimate the latent source \cite{owen2012performance,sekihara2015bayesian,wipf2009unified}.

Sparse Bayesian learning (SBL) provides a principled empirical-Bayes framework for M/EEG source imaging. Rather than fixing the source regularization a priori, SBL models source and noise coefficients with Gaussian priors whose variance hyperparameters are estimated directly from the measurements. In this way, sparsity promotion, hyperparameter adaptation, and posterior inference are handled within a single probabilistic model \cite{hashemi2021unification,owen2012performance,tipping2001sparse,wipf2009unified,cai2021robust}.

In classical SBL for M/EEG inverse modeling, the source and noise hyperparameters are estimated by minimizing a nonconvex Type-II marginal-likelihood objective through iterative update rules based on auxiliary objective functions, majorization, or fixed-point reformulations. The three commonly used update families in this setting are expectation-maximization, MacKay-type, and convex-bounding update rules \cite{sekihara2015bayesian}. Although all three can be used for hyperparameter learning, we adopt convex-bounding as the reference classical solver because, in the M/EEG literature, it is typically regarded as one of the most computationally efficient and fastest-converging alternatives, while also providing simple explicit update formulas that are especially convenient for the deep unfolding algorithms \cite{cai2021robust,hashemi2021unification,wipf2009unified}.

At the same time, recent deep learning approaches have shown that data-driven inference can substantially accelerate M/EEG brain source imaging and improve empirical reconstruction performance \cite{hecker2021convdip,liang2023sbl_dnn,morik2024physics3d,pantazis2021deepmeg,sun2022deepsif}. These methods span supervised source-imaging networks, hybrid combinations of Bayesian modeling and neural networks, and physically informed initialization strategies. However, these approaches generally use neural networks to predict source activity directly, to augment the inverse mapping, or to learn auxiliary representations, rather than treating the Type-II evidence-maximization updates themselves as the trainable object. Deep unfolding provides a natural motivation for this formulation, since it converts an iterative solver into a trainable layered architecture whose layers remain tied to the update structure of the original algorithm \cite{gregor2010learning,hershey2014deep,monga2021algorithm}. This is especially relevant for the Type-II Bayesian learning setting, because the source and noise hyperparameters explicitly control sparsity, data-fit weighting, and the evolution of the reconstruction. 

We therefore learn structured corrections to the joint source and noise hyperparameter dynamics, so that each layer remains anchored to the Bayesian solver while the finite-depth inference process is optimized for empirical reconstruction performance. Starting from a log-domain reparameterization of the classical positive hyperparameter updates, we enrich this baseline through a sequence of learned corrections, from bias-only adaptation to residual multi-layer perceptron (MLP) refinement and an attention-augmented residual variant. 
To the best of our knowledge, this is the first M/EEG source-imaging framework that unfolds and learns structured corrections to the \emph{joint Type-II source and noise hyperparameter update mechanism}.



\paragraph{Contributions.}
Our contributions are fourfold. First, we formulate joint Type-II Bayesian learning of source-wise diagonal prior variances and sensor-wise diagonal noise variances within an unfolded framework for M/EEG source imaging. Second, we derive structure-preserving log-domain neural parameterizations of the classical convex-bounding updates, including a baseline parameterization that exactly reproduces the original solver at initialization. Third, we introduce progressively more expressive learned corrections to this baseline, namely \textit{bias-only}, \textit{residual MLP}, and \textit{attention-augmented} residual variants. Fourth, we evaluate these variants in terms of reconstruction accuracy and convergence behavior, and show that residual-based refinements substantially outperform the baseline parameterization while retaining the interpretability of the original Bayesian algorithm.

\section{Related Work}

Classical Type-II Bayesian methods form a central model-based framework for electromagnetic source imaging. In M/EEG, they are commonly realized through $\gamma$-MAP or Champagne algorithms, which estimate source variance hyperparameters by marginal-likelihood maximization (e.g., Type-II ML) \cite{wipf2007analysis,wipf2009unified,wipf2010iterative}. Classical update schemes for this objective, including expectation-maximization (EM), MacKay-type, and convex-bounding updates, have been unified under a common majorization-minimization viewpoint \cite{hashemi2021unification}.

The same Type-II framework can also be extended to likelihood-noise hyperparameters. When the diagonal noise variances are also learned from the same marginal-likelihood principle, this yields the adaptive-noise extension with joint estimation of $(\boldsymbol{\gamma},\boldsymbol{\lambda})$ \cite{cai2021robust,wipf2009unified,wipf2010iterative}. 
Broader Type-II extensions have considered richer noise structures, including full noise covariance and spatio-temporal source-noise covariance models \cite{hashemi2021spatiotemporal, hashemi2024fullstructure}. Here, we focus on the source-wise and sensor-wise diagonal formulation because it yields explicit scalar source and noise hyperparameter updates that can be unfolded into separate source and noise branches.

Recent deep learning methods provide fast alternatives to classical source imaging, including supervised, hybrid Bayesian-neural, and physically informed approaches \cite{hecker2021convdip,liang2023sbl_dnn,morik2024physics3d,pantazis2021deepmeg,sun2022deepsif}. However, these methods generally do not retain the explicit Type-II hyperparameter-update mechanism itself. Deep unfolding offers a model-based alternative by converting iterative solvers into trainable layered architectures while preserving their algorithmic structure \cite{gregor2010learning,hershey2014deep,monga2021algorithm}. \rebuttal{Prior work has successfully learned covariance maps, parameters, or unfolded SBL blocks for generic sparse recovery, channel estimation, and direction-of-arrival estimation \cite{peter2019learnedsbl,hu2022ddpg,gao2023ampsbl,gong2023sblnet}. The concurrent method of \citet{balaji2026deepunfolding} learns source-side structure with known scalar noise.}  Our work follows this direction, but targets the joint source-prior and sensor-noise update dynamics of a classical convex-bounding solver. More broadly, the same correction-learning principle is relevant to Type-II sparse Bayesian learning, automatic relevance determination, and relevance-vector-machine models, where evidence maximization is used to update relevance or variance hyperparameters \cite{tipping2001sparse,wipf2007newview,wipf2004sparse}.

\section{Background}
\label{sec:background}

\subsection{Sparse Bayesian learning}
Let $\mathbf{Y}=[\mathbf{y}(1),\dots,\mathbf{y}(T)]\in\mathbb{R}^{M\times T}$ denote the sensor measurements and $\mathbf{X}=[\mathbf{x}(1),\dots,\mathbf{x}(T)]\in\mathbb{R}^{N\times T}$ denote the unknown source activities over $T$ time instants, and let $\mathbf{L}\in\mathbb{R}^{M\times N}$ be the lead-field matrix. We consider the linear forward model
\begin{equation}
\mathbf{y}(t)=\mathbf{L}\mathbf{x}(t)+\boldsymbol{\varepsilon}(t), \quad t=1,\dots,T.
\end{equation}
For each source index $n=1,\dots,N$ and sensor index $m=1,\dots,M$, we assume the scalar source and noise variables satisfy
independently across source indices, sensor indices, and time instants $t=1,\dots, T$ 
\begin{equation}
\mathbf{x}(t)\sim\mathcal{N}(\mathbf{0},\boldsymbol{\Gamma}), \qquad \boldsymbol{\varepsilon}(t)\sim\mathcal{N}(\mathbf{0},\boldsymbol{\Lambda}),
\end{equation}
with source-wise and sensor-wise diagonal covariance matrices
\begin{equation*}
\boldsymbol{\Gamma}=\operatorname{diag}(\gamma_1,\dots,\gamma_N), \qquad \boldsymbol{\Lambda}=\operatorname{diag}(\lambda_1,\dots,\lambda_M).
\end{equation*}
The standard homoscedastic noise model $\sigma^2\mathbf{I}$ is recovered as the special case $\lambda_1=\ldots=\lambda_M\coloneqq\sigma^2$.

Under the temporal independence assumptions on the source prior and sensor noise, the prior and likelihood factorize over time as
\begin{equation*}
p(\mathbf{X}\mid\boldsymbol{\gamma})=\prod_{t=1}^T \mathcal{N}\!\bigl(\mathbf{x}(t)\mid \mathbf{0},\boldsymbol{\Gamma}\bigr), \qquad
p(\mathbf{Y}\mid\mathbf{X},\boldsymbol{\lambda})=\prod_{t=1}^T \mathcal{N}\!\bigl(\mathbf{y}(t)\mid \mathbf{L}\mathbf{x}(t),\boldsymbol{\Lambda}\bigr).
\end{equation*}
Consequently, the posterior also factorizes across time,
\begin{equation}
p(\mathbf{X}\mid\mathbf{Y},\boldsymbol{\gamma},\boldsymbol{\lambda})
=
\prod_{t=1}^T \mathcal{N}\!\bigl(\bar{\mathbf{x}}(t),\boldsymbol{\Sigma}_\mathbf{x}\bigr),
\end{equation}
with posterior mean $\bar{\mathbf{x}}(t)$ and covariance $\boldsymbol{\Sigma}_\mathbf{x}$ depending on the marginal data or 
statistical covariance $\boldsymbol{\Sigma}_\mathbf{y}$
\begin{equation}
\bar{\mathbf{x}}(t)=\boldsymbol{\Gamma}\mathbf{L}^\top\boldsymbol{\Sigma}_\mathbf{y}^{-1}\mathbf{y}(t), \qquad
\boldsymbol{\Sigma}_\mathbf{x}=\boldsymbol{\Gamma}-\boldsymbol{\Gamma}\mathbf{L}^\top\boldsymbol{\Sigma}_\mathbf{y}^{-1}\mathbf{L}\boldsymbol{\Gamma}, \qquad \boldsymbol{\Sigma}_\mathbf{y}=\mathbf{L}\boldsymbol{\Gamma}\mathbf{L}^\top+\boldsymbol{\Lambda}.
\label{eq:posterior_summaries}
\end{equation}
These posterior summaries are the central quantities used by both classical sparse Bayesian solvers, namely: $\gamma$-MAP or Champagne algorithms \cite{wipf2007analysis,wipf2009unified,wipf2010iterative} and our unfolded network proposed later.

\subsection{Type-II hyperparameter learning}

SBL for electromagnetic brain imaging is formulated within the Type-II Bayesian framework, in which model (e.g., source and noise) hyperparameters are estimated by maximizing the marginal likelihood of the measurements \cite{sekihara2015bayesian,wipf2009unified,wipf2010iterative}. In the present setting, these hyperparameters are the source-wise variance parameters $\boldsymbol{\gamma}$ and, in the adaptive noise learning case, the sensor-wise noise variance parameters $\boldsymbol{\lambda}$. 
Accordingly, the hyperparameters are estimated by minimizing the negative marginal log-likelihood
\begin{equation}
\mathcal{L}_{\mathrm{II}}(\boldsymbol{\gamma},\boldsymbol{\lambda})
=
\log\bigl|\boldsymbol{\Sigma}_\mathbf{y}\bigr|
+
\frac{1}{T}\sum_{t=1}^T \mathbf{y}(t)^\top \boldsymbol{\Sigma}_\mathbf{y}^{-1}\mathbf{y}(t).
\label{eq:typeII_loss}
\end{equation}
Since \eqref{eq:typeII_loss} is nonconvex in the hyperparameters, classical $\gamma$-MAP methods optimize it through iterative surrogate schemes such as expectation-maximization (EM), MacKay-type, and convex-bounding update rules \cite{hashemi2021unification,wipf2009unified,wipf2010iterative}. In particular, the convex-bounding approach replaces the original nonconvex Type-II objective by a convex majorizing surrogate with respect to the underlying hyperparameters at each iteration, yielding simple closed-form update rules and extending naturally from source-prior learning to adaptive sensor-noise learning \cite{cai2021robust,hashemi2021unification}. In this paper, we take this convex-bounding route for both hyperparameter blocks as the classical \emph{baseline} to be mimicked and refined, while the \emph{correction learning theory} generalizes to different update rules.

\subsection{Classical convex-bounding baseline}

The classical convex-bounding baseline performs alternating hyperparameter learning over source and noise variances. At iteration $k=1,2,\dots,K$, one first computes the posterior mean using the current hyperparameter values $(\boldsymbol{\Gamma}^{(k)},\boldsymbol{\Lambda}^{(k)})$, then updates the source-wise variances $\gamma_n$ with the current noise variances fixed, and finally updates the sensor-wise noise variances $\lambda_m$ using the updated source variances. Thus, both hyperparameter blocks are refined \emph{alternately} within the same Type-II loss scheme. Throughout, the same index $k$ denotes a classical solver iteration and its corresponding unfolded layer, with $(k)$ used for iteration indices and $k$ for layer indices.

For the \emph{source branch}, define at unfolded layer $k$
\begin{equation*}
e_n^{k}=\frac{1}{T}\sum\limits_{t=1}^T \bigl(\bar{x}_n^{k}(t)\bigr)^2, \qquad
z_n^{k}=\mathbf{l}_n^\top (\boldsymbol{\Sigma}_y^{k})^{-1}\mathbf{l}_n, \quad n=1,\dots,N,
\end{equation*}
and for the \textit{noise branch} define
\begin{equation*}
d_m^{k}=\frac{1}{T}\sum\limits_{t=1}^T \Bigl(y_m(t)-\mathbf{l}_{m:}\bar{\mathbf{x}}^{k}(t)\Bigr)^2, \qquad
q_m^{k}=\bigl[(\boldsymbol{\Sigma}_y^{k})^{-1}\bigr]_{mm}, \quad m=1,\dots,M.
\end{equation*}
where $\mathbf{l}_n$ and $\mathbf{l}_{m:}$ denote the $n$th column and the $m$th row of lead field matrix $\mathbf{L}$, respectively. \cite{cai2021robust,hashemi2021unification,wipf2009unified,wipf2010iterative}.
 The \emph{source-wise} and \emph{sensor-wise} convex-bounding updates are defined via
\begin{equation}
\gamma_n^{(k+1)}
=
\sqrt{
\frac{
e_n^{k}
}{
z_n^{k}
}
},  \quad n=1,\dots,N,
\qquad
\lambda_m^{(k+1)}
=
\sqrt{
\frac{
d_m^{k}
}{
q_m^{k}
}
}, \quad m=1,\dots,M.
\label{eq:cb_updates}
\end{equation}
Hence, both hyperparameter blocks follow the same square-root ratio pattern, with an empirical energy term in the numerator and a curvature term in the denominator. These alternating update equations \ref{eq:cb_updates} define the \emph{exact model-based convex-bounding baseline} used throughout the remainder of the paper.

\section{Methods}
\label{sec:method}

\subsection{Stable log-domain feature representation}

We recast the convex-bounding updates \eqref{eq:cb_updates} into a
numerically stabilized \emph{log-domain} form to enforce variance
positivity and transform the problem into \emph{additive correction learning}. Operating in this logarithmic space compresses the dynamic range of the statistics, thereby ensuring numerical stability and stabilizing the neural network training. More specifically, the learned modules receive the log-transformed convex-bounding statistics $(u^k,v^k)$ for the \textit{source branch} and $(r^k,s^k)$ for the \textit{noise branch}, derived from $(e^k,z^k)$ and $(d^k,q^k)$ below, while the \textit{analytical convex-bounding update} itself remains explicit. 

To ensure that the logarithms are well-defined even when their nonnegative arguments vanish, we introduce small positive stabilization constants $\varepsilon_\gamma$ and $\varepsilon_\lambda$ and define the stabilized log-features as
\begin{equation*}
\begin{aligned}
u_n^{k} &= \log(e_n^{k}+\varepsilon_\gamma), \quad
v_n^{k} = \log(z_n^{k}+\varepsilon_\gamma),
& n&=1,\dots,N,\\[4pt]
r_m^{k} &= \log(d_m^{k}+\varepsilon_\lambda), \quad
s_m^{k} = \log(q_m^{k}+\varepsilon_\lambda),
& m&=1,\dots,M.
\end{aligned}
\end{equation*}
Let $\boldsymbol{\ell}_\gamma^{k}=\log \boldsymbol{\gamma}^{k}$ and $\boldsymbol{\ell}_\lambda^{k}=\log \boldsymbol{\lambda}^{k}$. Then the classical convex-bounding baseline becomes
\begin{equation}
\ell_{\gamma,n}^{k+1}=\tfrac{1}{2}u_n^{k}-\tfrac{1}{2}v_n^{k},\quad n=1,\dots,N, \qquad
\ell_{\lambda,m}^{k+1}=\tfrac{1}{2}r_m^{k}-\tfrac{1}{2}s_m^{k},\quad m=1,\dots,M.
\label{eq:log_cb_scalar}
\end{equation}
followed by $\gamma_n^{k+1}=\exp(\ell_{\gamma,n}^{k+1})$ and $\lambda_m^{k+1}=\exp(\ell_{\lambda,m}^{k+1})$. Thus, the original \emph{multiplicative square-root updates} are converted into \emph{additive log-domain updates} built from the classical convex-bounding statistics $(e^k,z^k,d^k,q^k)$.

\subsection{Correction-learning update families}

Our central idea is to refine the classical convex-bounding updates
through \emph{correction learning}, retaining their analytical structure
while learning additive corrections in the log-domain to improve
finite-depth reconstruction. This follows the residual-learning
principle of expressing a trainable mapping as a reference mapping
plus a learned correction \cite{he2016deep}, and is related to
solver-in-the-loop approaches that train corrections through numerical
solver dynamics \cite{um2020solver}. Specifically, we retain the
analytical log-ratio structure in \eqref{eq:log_cb_scalar}, including
the fixed coefficients $+\tfrac{1}{2}$ and $-\tfrac{1}{2}$, and learn
only the correction terms for the source and noise hyperparameter
updates. We denote the trainable source-branch and noise-branch
parameters by $\theta$ and $\phi$, respectively.
This yields three increasingly expressive variants.

\paragraph{Bias-only correction (Bias CB).}
The simplest variant adds learnable bias vectors
$\mathbf{c}_\gamma^{k} \in\mathbb{R}^{N}$
and
$\mathbf{c}_\lambda^{k}\in\mathbb{R}^{M}$
at each unfolded layer:
\begin{equation}
\boldsymbol{\ell}_\gamma^{k+1}=\tfrac{1}{2}\mathbf{u}^{k}-\tfrac{1}{2}\mathbf{v}^{k}+\mathbf{c}_\gamma^{k}, \qquad
\boldsymbol{\ell}_\lambda^{k+1}=\tfrac{1}{2}\mathbf{r}^{k}-\tfrac{1}{2}\mathbf{s}^{k}+\mathbf{c}_\lambda^{k}.
\label{eq:bias_only}
\end{equation}
Thus, in the bias-only variant, the correction is specific to both
the unfolded layer and the individual source or sensor while being
input agnostic. 

\paragraph{Residual MLP correction (Deep CB).}
The second variant makes the correction input-dependent through two pointwise MLPs, one for the \emph{source branch} and one for the \emph{noise branch}:
\begin{equation}
\begin{aligned}
\ell_{\gamma,n}^{k+1}
&=
\tfrac{1}{2}u_n^{k}-\tfrac{1}{2}v_n^{k}
+
f_\gamma^\theta\!\bigl(u_n^{k},v_n^{k},\log(\gamma_n^{k}+\varepsilon_\gamma)\bigr), \\[1.0ex]
\ell_{\lambda,m}^{k+1}
&=
\tfrac{1}{2}r_m^{k}-\tfrac{1}{2}s_m^{k}
+
f_\lambda^\phi\!\bigl(r_m^{k},s_m^{k},\log(\lambda_m^{k}+\varepsilon_\lambda)\bigr).
\end{aligned}
\label{eq:mlp_system}
\end{equation}
Here $f_\gamma^\theta:\mathbb{R}^3\to\mathbb{R}$ and $f_\lambda^\phi:\mathbb{R}^3\to\mathbb{R}$ are small pointwise multilayer perceptrons. Each function is shared across all sources or all sensors, respectively, and is applied independently to the local feature vector of each source or sensor. Therefore, unlike the bias-only correction, the MLP correction does not introduce a separate trainable parameter for every source or sensor, but generalizes to new source spaces and sensor layouts. 
Consequently, for large $N$ or $M$, the shared pointwise MLP variant can have fewer trainable parameters than the bias-only model, even though the correction rule is more expressive. Detailed architectural choices for these MLPs are deferred to Appendix~\ref{app:bias_mlp}.

\paragraph{Attention-augmented residual correction (Deep Attn. CB).}
The third variant keeps the same residual structure but enriches the \emph{source-side} correction by allowing interactions across sources.
\begin{equation}
\mathbf{H}_\gamma^{k}=
\bigl[
(u_1^{k},v_1^{k},\log(\gamma_1^{k}+\varepsilon_\gamma)),\dots,
(u_N^{k},v_N^{k},\log(\gamma_N^{k}+\varepsilon_\gamma))
\bigr]^\top \in \mathbb{R}^{N\times 3}
\end{equation}
We define the attention module as a shared sequence-to-vector map
$\mathcal{A}_\gamma^\theta:\mathbb{R}^{N\times 3}\to\mathbb{R}^{N}$, whose $n$th output component gives the \emph{residual correction} for source $n$. Thus, the module is not source-specific in its parameters, but its outputs are source-specific because each correction depends on the local feature vector of that source together with the feature vectors of the other sources.
We then replace the pointwise source correction by
\begin{equation}
\boldsymbol{\ell}_\gamma^{k+1}
=
\tfrac{1}{2}\mathbf{u}^{k}-\tfrac{1}{2}\mathbf{v}^{k}
+
\mathcal{A}_\gamma^\theta(\mathbf{H}_\gamma^{k}),
\label{eq:attn_gamma}
\end{equation}
while the sensor-noise branch remains in the residual MLP form \eqref{eq:mlp_system}. In this way, the source correction is no longer purely pointwise: the update for each source depends on the global configuration of source-wise convex-bounding features. Since no explicit anatomical coordinates are included in $\mathbf{H}_\gamma^k$, this module learns global source-wise context rather than explicit geometric spatial context. Detailed attention-module design is deferred to Appendix~\ref{app:attention}.

\paragraph{Architectural preservation versus optimization guarantees.}
\rebuttal{
In this work, structure preservation is \emph{architectural} rather than \emph{optimization-theoretic}. The learned updates retain the explicit CB log-ratio term, enforce positivity through exponentiation, and reduce exactly to the classical CB update when the \textit{residual correction} vanishes. This exact equivalence therefore holds at zero correction, including at the proposed initialization, but not generally after learning. Once a nonzero correction is applied, the resulting iterate is no longer the update obtained by minimizing the classical CB majorizing surrogate. Consequently, the majorization-minimization descent argument cannot be invoked, and neither monotone decrease of the Type-II objective nor the solution-level convergence properties of the classical solver are claimed. Thus, \emph{structure-preserving} refers to retaining the analytical form of the baseline update and does not imply that the learned corrections preserve its descent or convergence guarantees. Appendix~\ref{app:typeii_objective} empirically illustrates the resulting non-monotone behavior.}

\paragraph{Exact classical initialization.}
All three variants are initialized so that the learned correction is identically zero at the start of training. For the bias-only model, this means $\mathbf{c}_\gamma^{k}=\mathbf{0}$ and $\mathbf{c}_\lambda^{k}=\mathbf{0}$. For the MLP and attention-based models, the final output layers are initialized to zero, so that the correction term vanishes initially. Consequently, every variant exactly recovers the \emph{baseline solver (e.g., classical convex-bounding update rule)} before learning begins. Exact recovery refers to the baseline implementation with the same numerical stabilization and sparsification settings. \rebuttal{Appendix~\ref{app:zero_initialization} separates this structural property from optimization by comparing zero and standard learned-layer initialization. The comparison shows comparable final reconstruction accuracy under both initialization strategies, suggesting that zero initialization provides an interpretable model-based starting point but is not essential for the reported accuracy. Standard random initialization retains the log-domain parameterization and positivity. Initialization may nevertheless affect early training dynamics and variability across runs.}

\subsection{Alternating unfolded architecture}

Each unfolded layer mirrors one iteration of the classical alternating convex-bounding solver. Starting from $(\boldsymbol{\Gamma}^{k},\boldsymbol{\Lambda}^{k})$, layer $k$ first computes the posterior state $(\bar{\mathbf{x}}^{k},\boldsymbol{\Sigma}_\mathbf{x}^{k})$. From this state, the source statistics $(e^k,z^k)$ are formed, and the source variances are updated through one of the correction-learning rules above. In the source-only $\gamma$-MAP setting, the noise covariance $\boldsymbol{\Lambda}^{k}$ is kept fixed. In the joint source-noise setting, using the updated source covariance $\boldsymbol{\Gamma}^{k+1}$ together with the previous noise covariance $\boldsymbol{\Lambda}^{k}$, an intermediate posterior state is recomputed. The noise statistics $(d^k,q^k)$ are computed from this intermediate
posterior state and used to update the sensor-noise variances. Finally, the posterior state is recomputed with $(\boldsymbol{\Gamma}^{k+1},\boldsymbol{\Lambda}^{k})$ in the source-only case or with $(\boldsymbol{\Gamma}^{k+1},\boldsymbol{\Lambda}^{k+1})$ in the joint case and passed to layer $k+1$. In this way, the unfolded network preserves the \textit{alternating} state evolution of the classical solver, while localizing learning to explicit correction terms. Algorithm~\ref{alg:unfolded_layer} summarizes one unfolded layer, with the noise-update step included only for the joint source-noise variant.

\begin{figure*}[!t]
    \centering
    \includegraphics[width=\textwidth]{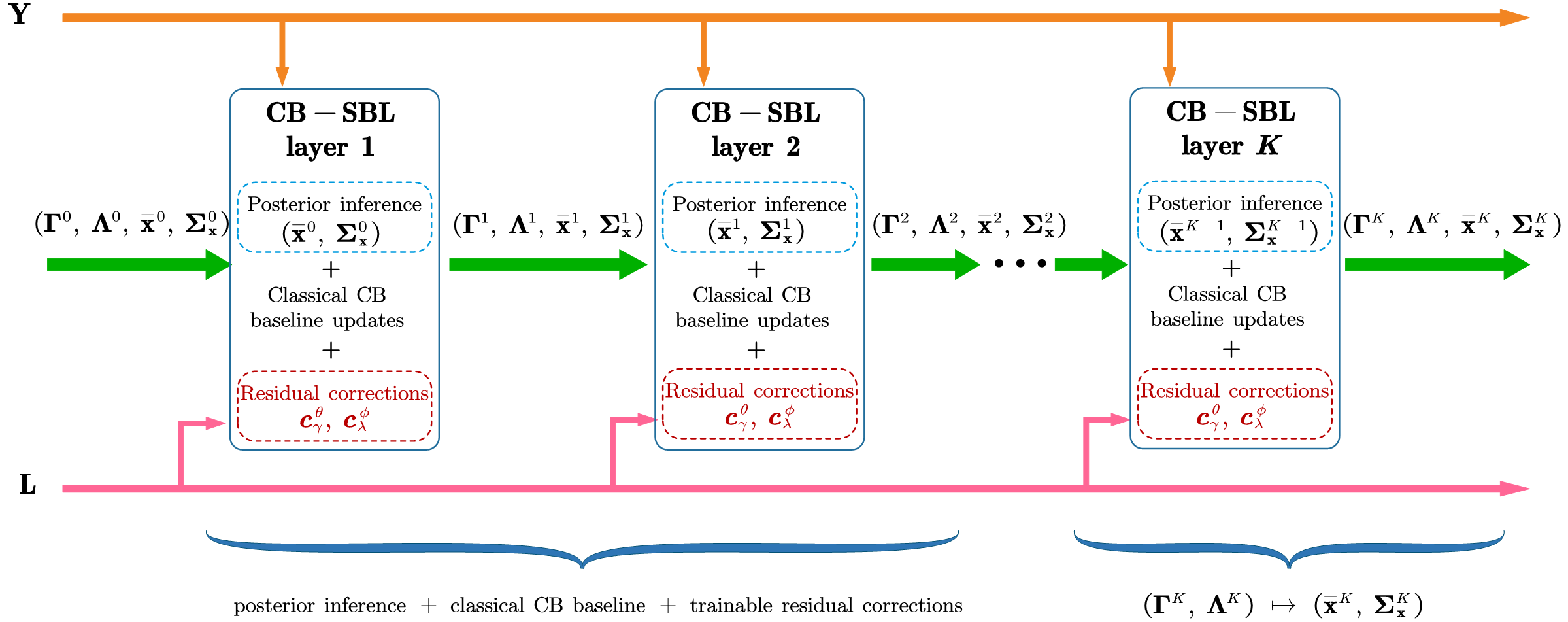}
    \caption{Structure-preserving unfolded architecture for \emph{residual correction learning}. Each trainable CB-SBL layer receives the data $\mathbf{Y}$ and lead field $\mathbf{L}$, performs posterior inference, applies the classical convex-bounding baseline updates, and adds learned residual corrections for the source and, in the joint case, noise branches. Thus, each unfolded layer mimics one step of the classical solver while remaining trainable through correction learning. The final layer outputs the posterior mean and covariance induced by the final hyperparameter estimates.}
    \label{fig:unfolded_architecture}
\end{figure*}

Figure~\ref{fig:unfolded_architecture} illustrates this construction: each trainable CB-SBL layer combines posterior inference, the classical convex-bounding baseline, and learned residual corrections, whereas the final layer remains purely model-based and untrainable. To stabilize optimization and avoid the pitfalls of simultaneously training all layers, we employ an iterative, layer-wise training strategy \cite{monga2021algorithm}. Additional implementation details for the correction modules and training regularization are deferred to Appendix~\ref{app:details}.

\begin{algorithm}[!t]
\caption{One unfolded layer for source-only or joint source-noise correction learning}
\label{alg:unfolded_layer}
\begin{algorithmic}[1]
\Require $(\boldsymbol{\Gamma}^{k},\boldsymbol{\Lambda}^{k})$, data $\mathbf{Y}$, lead field $\mathbf{L}$, mode $\in\{\text{source-only},\text{joint}\}$
\State Compute $\boldsymbol{\Sigma}_\mathbf{y}^{k}=\mathbf{L}\boldsymbol{\Gamma}^{k}\mathbf{L}^\top+\boldsymbol{\Lambda}^{k}$ and posterior mean $\bar{\mathbf{x}}^{k}(t)$
\State Form source statistics $(e^k,z^k)$ and log-features $(\mathbf{u}^{k},\mathbf{v}^{k})$
\State Update $\boldsymbol{\Gamma}^{k+1}$ using \eqref{eq:bias_only}, \eqref{eq:mlp_system}, or \eqref{eq:attn_gamma}
\If{mode is joint}
    \State Recompute the intermediate posterior state with $(\boldsymbol{\Gamma}^{k+1},\boldsymbol{\Lambda}^{k})$
    \State Form noise statistics $(d^k,q^k)$ and log-features $(\mathbf{r}^{k},\mathbf{s}^{k})$
    \State Update $\boldsymbol{\Lambda}^{k+1}$ using \eqref{eq:bias_only} or \eqref{eq:mlp_system}
    \State Recompute the posterior state with $(\boldsymbol{\Gamma}^{k+1},\boldsymbol{\Lambda}^{k+1})$
\Else
    \State Set $\boldsymbol{\Lambda}^{k+1}=\boldsymbol{\Lambda}^{k}$ and recompute the posterior state with $(\boldsymbol{\Gamma}^{k+1},\boldsymbol{\Lambda}^{k})$
\EndIf
\State \Return $(\boldsymbol{\Gamma}^{k+1},\boldsymbol{\Lambda}^{k+1},\bar{\mathbf{x}}^{k+1},\boldsymbol{\Sigma}_\mathbf{x}^{k+1})$
\end{algorithmic}
\end{algorithm}

\subsection{End-to-end training}
Let $\Theta$ denote the trainable parameters of the chosen correction modules. For an unfolded network with $K$ layers, let $\bar{\mathbf{x}}_\Theta^{K}(t)$ denote the final posterior mean estimate. 
For a training pair $(\mathbf{Y},\mathbf{X}^\star)$, we define
the reconstruction loss
\begin{equation}
\mathcal{J}(\Theta)
=
\frac{1}{T}\sum\limits_{t=1}^T
\left\|
\bar{\mathbf{x}}_\Theta^{K}(t)-\mathbf{x}^{\star}(t)
\right\|_2^2.
\end{equation}
During training, this loss is averaged over the samples in each mini-batch.
Hence, \emph{source-side} and \emph{noise-side} corrections are learned jointly through a common end-to-end objective, while every layer remains anchored to the classical convex-bounding dynamics. To ensure that early layers still produce good estimates, we add a \emph{stochastic deep-supervision regularizer} acting on intermediate layers as described in Appendix~\ref{app:deep_supervision}.


\section{Experiments}
\label{sec:experiments}
Because \textit{in vivo} cortical ground truth is fundamentally unavailable, we evaluate our approach on a rigorously controlled synthetic M/EEG benchmark. We simulate distributed, frequency-band-filtered neural sources projected through a realistic \texttt{fsaverage} forward model and deliberately corrupted by severe sensor-wise heteroscedastic noise (full details in Appendix~\ref{app:synthetic_generation}). Our experiments are designed to answer three core questions: (1) Does iterative correction learning improve source reconstruction, even in the difficult joint ($\boldsymbol{\Gamma}$,$\boldsymbol{\Lambda}$) setting? (2) Does structure-preserving unrolling accelerate convergence?  (3) Is the learned iterative framework robust to different settings?

To systematically address these questions, we evaluate three progressively expressive variants of our architecture: \textbf{Bias-CB} (layer-wise scalar biases), \textbf{Deep-CB} (shared pointwise MLPs), and \textbf{Deep-Attn-CB} (augmenting the source MLP with spatial self-attention), both in the source-only setting ($\boldsymbol{\Gamma}$) and joint learning $(\boldsymbol{\Gamma},\boldsymbol{\Lambda})$. We benchmark these against \textbf{sLORETA} \cite{pascual2002standardized}, classical source-only \textbf{Convex Bounding ($\boldsymbol{\Gamma}$)} \cite{wipf2009unified}, classical joint \textbf{Convex Bounding ($\boldsymbol{\Gamma},\boldsymbol{\Lambda}$)} \cite{cai2021robust}, and \textbf{DeepSIF} \cite{sun2022deepsif} as a representative end-to-end deep solver. When learning only $\boldsymbol{\Gamma}$, we provide ground-truth noise statistics $\boldsymbol{\Lambda}_0$. We report four evaluation metrics: Earth Mover's Distance (EMD) and localization error (LE) for spatial localization, relative mean squared error (rMSE) for signal fidelity, and F1-score for support recovery (see Appendix~\ref{app:metrics} for definitions).

\subsection{General Performance}
\label{sec:visual_results}

\begin{figure}[!t]
    \centering
    \begin{subfigure}[b]{0.48\textwidth}
        \centering
        \includegraphics[width=\linewidth]{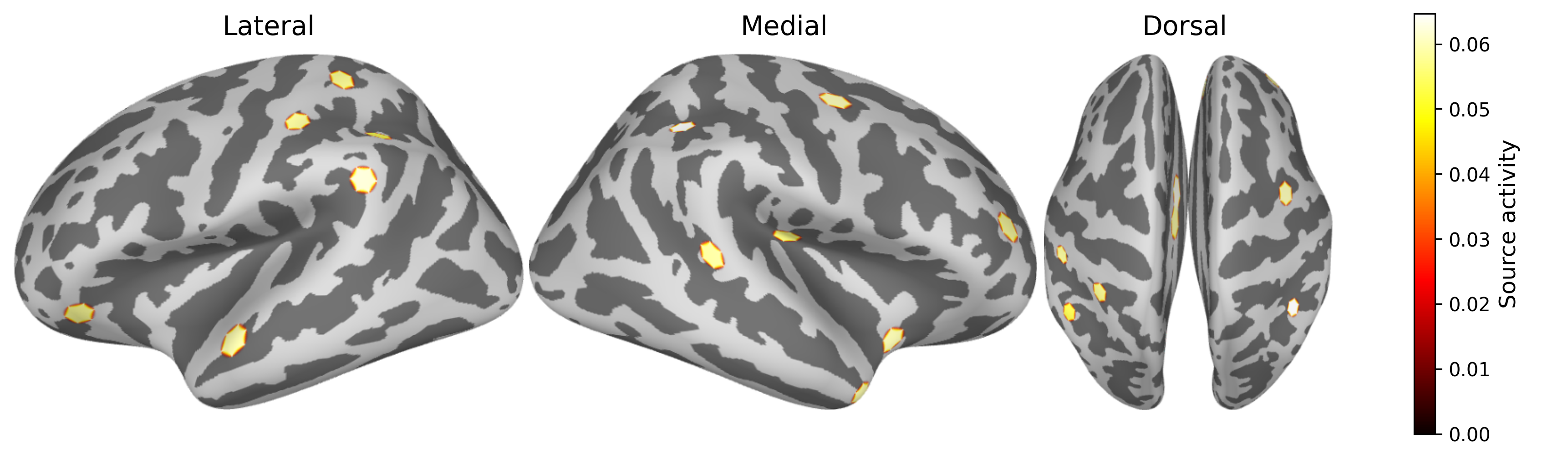} 
        \caption{Ground Truth (GT)}
        \label{fig:vis_gt}
    \end{subfigure}
    \hfill
    \begin{subfigure}[b]{0.48\textwidth}
        \centering
        \includegraphics[width=\linewidth]{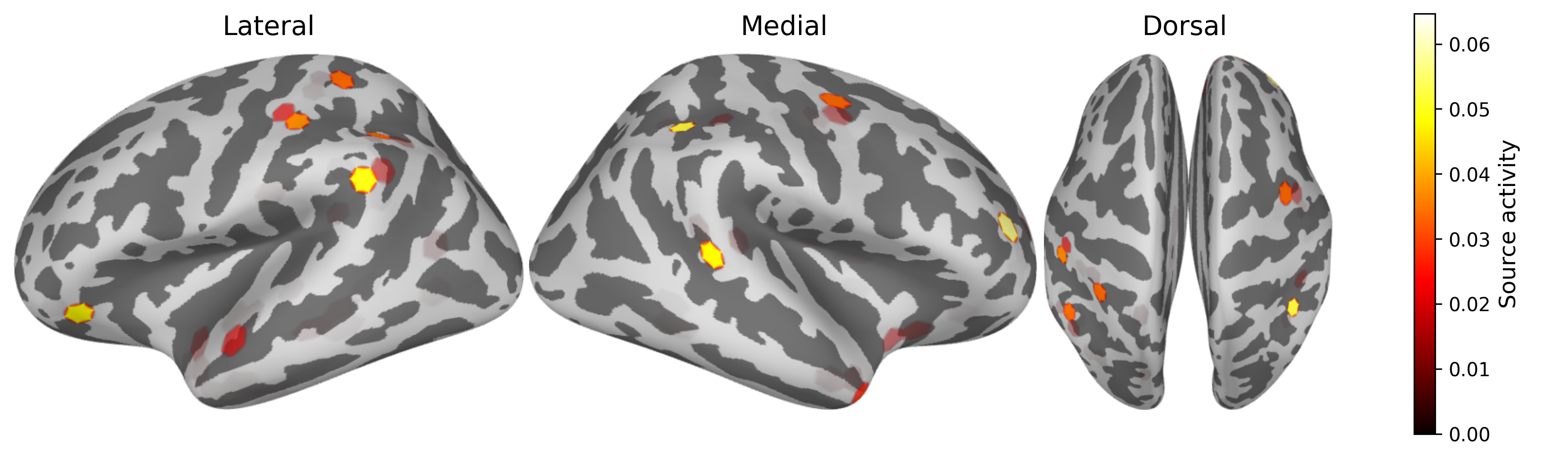}
        \caption{Convex Bounding}
        \label{fig:vis_cb}
    \end{subfigure}
    
    \vspace{0.4cm} 
    
    \begin{subfigure}[b]{0.48\textwidth}
        \centering
        \includegraphics[width=\linewidth]{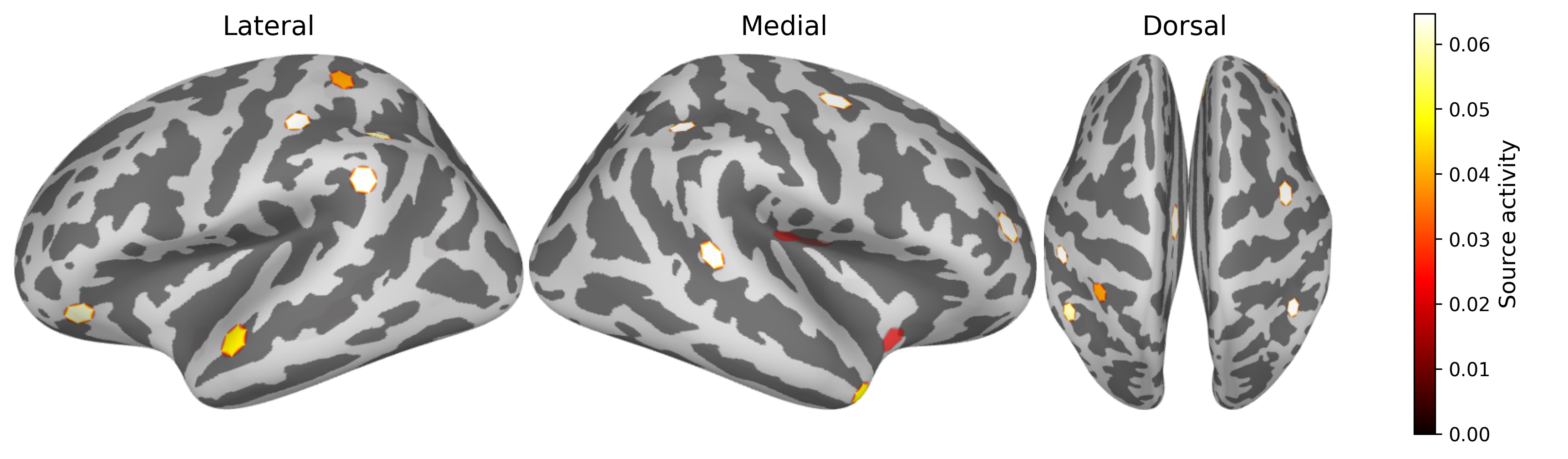}
        \caption{Deep CB}
        \label{fig:vis_deep_cb}
    \end{subfigure}
    \hfill
    \begin{subfigure}[b]{0.48\textwidth}
        \centering
        \includegraphics[width=\linewidth]{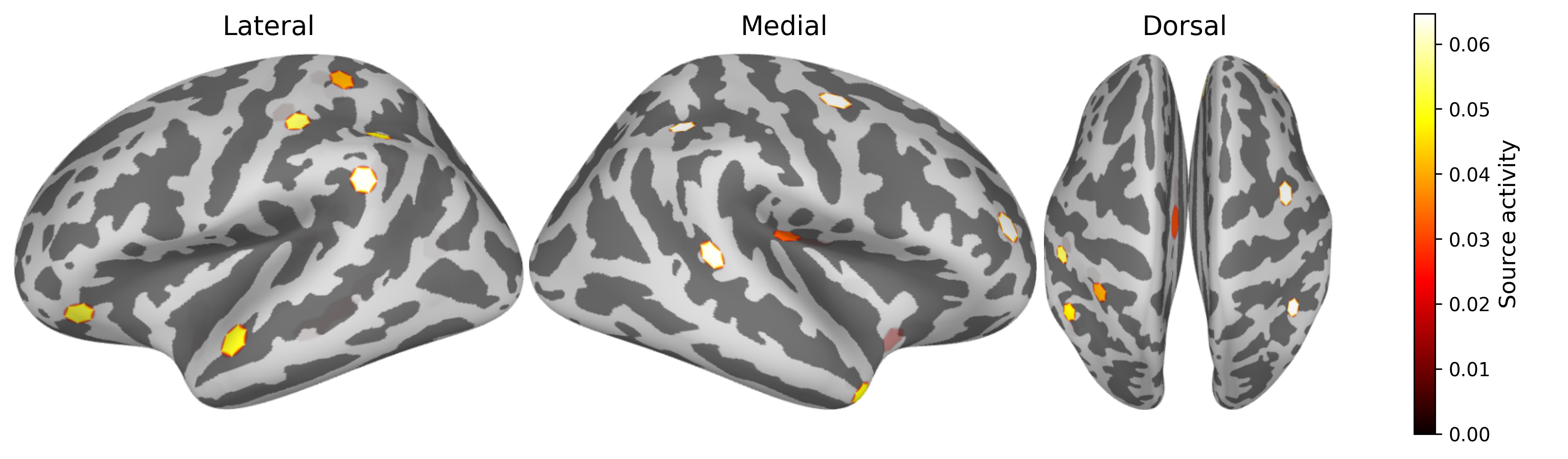}
        \caption{Deep Attn. CB}
        \label{fig:vis_deep_attn_cb}
    \end{subfigure}

    \vspace{0.25cm}
    \begin{subfigure}[b]{\linewidth}
        \centering
        \includegraphics[width=0.88\linewidth]{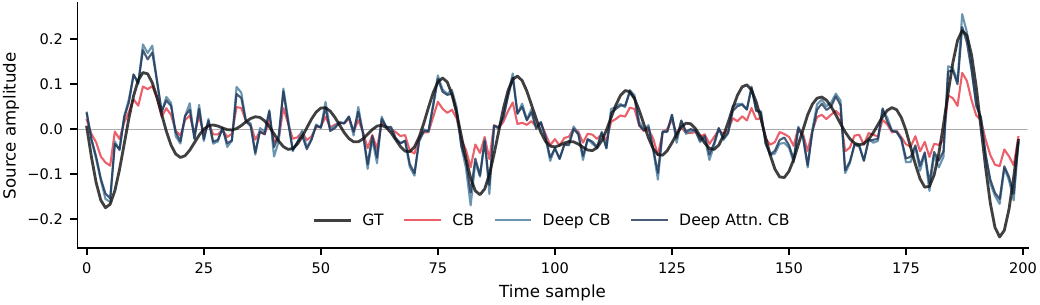}
        \caption{\rebuttal{Source-space time courses at an active ground-truth vertex}}
        \label{fig:vis_temporal}
    \end{subfigure}
    \caption{Qualitative comparison of cortical source activity reconstructions. The learned correction models \textbf{(c, d)} demonstrate superior spatial localization and sharper activity recovery compared to the baseline Convex Bounding \textbf{(b)}, more closely matching the Ground Truth \textbf{(a)} across lateral, medial, and dorsal views. \rebuttal{\textbf{(e)} Ground truth and estimates at the same active vertex show that the reconstructions also preserve time-resolved source dynamics.}}
    \label{fig:visual_results}
\end{figure}
To qualitatively evaluate localization performance, Figure~\ref{fig:visual_results} compares the reconstructed cortical source activity on the difficult Things Cap (Ico3) setting with SNR in $\mathcal{U}(5,20)$ with $25$ sources.  Both Deep CB and Deep Attn. CB capture the underlying source activity with higher fidelity, yielding sharper and more accurate active regions that align more closely with the ground truth than the classical convex bounding. \rebuttal{Panel~\ref{fig:vis_temporal} shows the temporal course for a source detected by all methods, where the learned variants fit the amplitude better.} For zero-shot real-world validation on the THINGS-EEG2 dataset \cite{gifford2022large}, the less complex Deep CB dominates as shown in Appendix~\ref{sec:realworlddata}.

\begin{figure}[t]
    \centering
    \vspace{-0.2cm}
    \begin{subfigure}{0.32\linewidth}
        \centering
        \includegraphics[width=\linewidth]{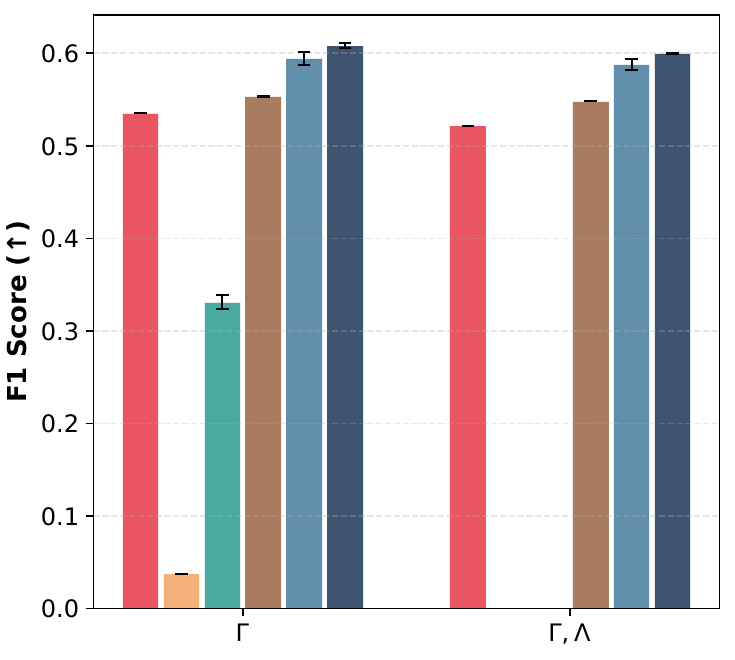}
    \end{subfigure}
    \hfill
    \begin{subfigure}{0.32\linewidth}
        \centering
        \includegraphics[width=\linewidth]{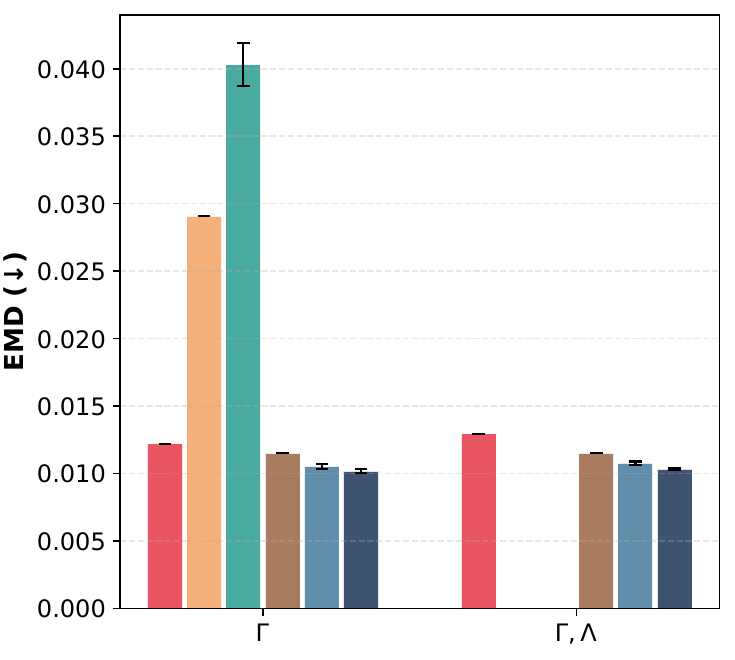}
    \end{subfigure}
    \hfill
    \begin{subfigure}{0.32\linewidth}
        \centering
        \includegraphics[width=\linewidth]{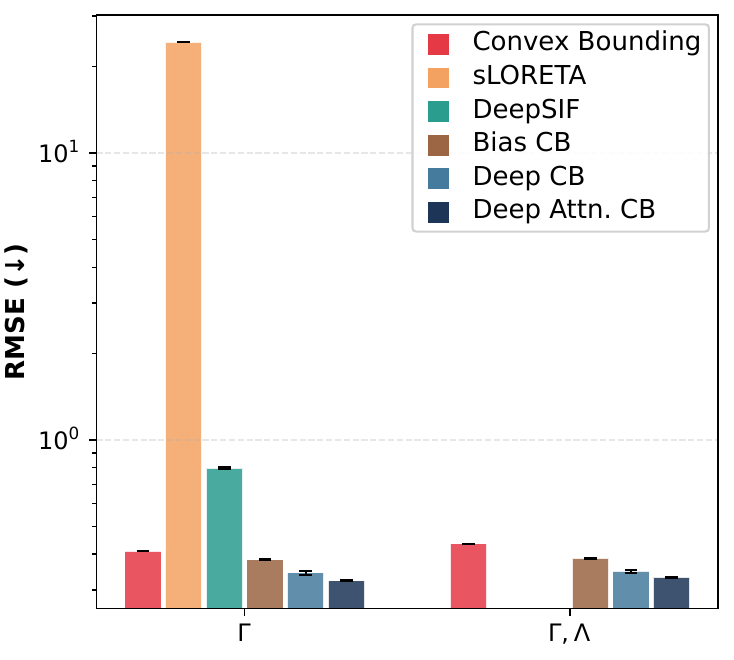}
    \end{subfigure}

    \caption{Comparison of overall performance metrics for both the ideal setting with ground-truth noise ($\boldsymbol{\Gamma}$) and the joint learning setting ($\boldsymbol{\Gamma}, \boldsymbol{\Lambda}$). Our approaches (Bias CB, Deep CB and Deep Attn. CB) consistently dominate across all metrics, with joint learning introducing only negligible performance drops. Note that the non-sparse baselines sLORETA and DeepSIF fail for our sparse setting.}
    \label{fig:overall_metrics}
\end{figure}

For quantitative evaluation, Figure~\ref{fig:overall_metrics} shows stepwise improvements progressing from the traditional Convex Bounding baseline, over the input-independent correction Bias CB, to the adaptive correction approaches (Deep CB and Deep Attn. CB). In contrast, dense methods like sLORETA and DeepSIF exhibit massive errors when attempting to localize sparse sources. The grouped bars confirm the stability of Convex Bounding also for unknown heteroscedastic noise as the joint setting ($\boldsymbol{\Gamma}, \boldsymbol{\Lambda}$) closely mirrors the performance of the models operating with ground-truth noise covariance ($\boldsymbol{\Gamma}$). Furthermore, small error bars computed over $5$ random seeds confirm stable performance independent of neural network initialization and training. More results on various validation settings with different sensor and source layouts can be found in Appendix~\ref{sec:general_performance}.

\subsection{Convergence Behavior and Importance of Regularization Loss}
\label{sec:convergence}
\begin{figure}[t]
    \centering
    \vspace{-0.2cm}
    \includegraphics[width=\linewidth]{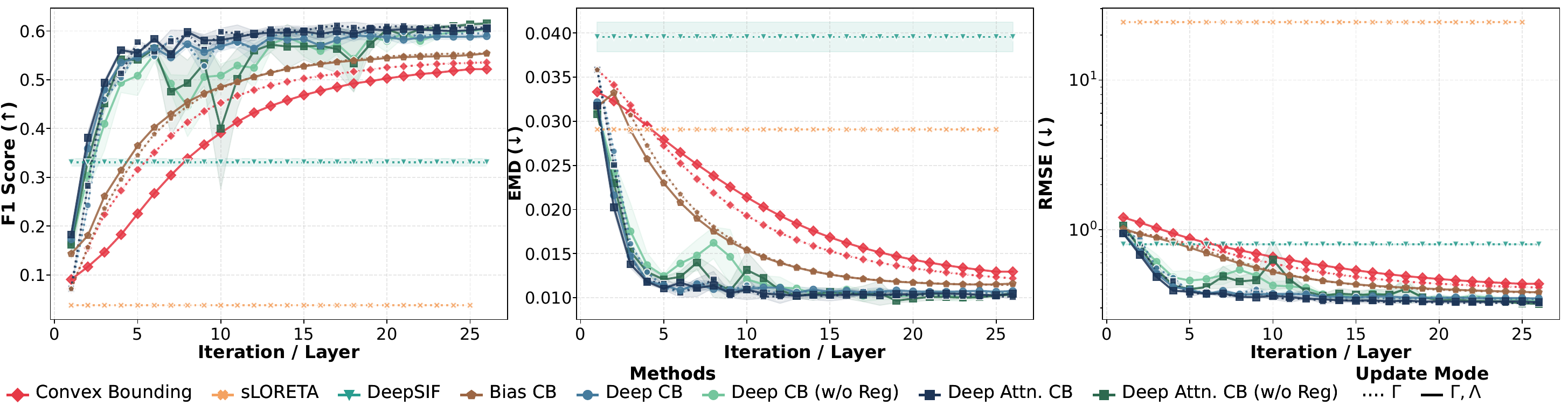}
    \caption{Test performance across intermediate algorithmic steps (network layers) highlights that deep attention architectures achieve faster convergence compared to traditional baselines, while reaching \rebuttal{better reconstruction performance}.
    }
    \label{fig:convergence_plots}
    \vspace{-1.5em}
\end{figure}

To empirically validate the stability and efficiency of our proposed architectures, we analyze the test-time performance across intermediate layers. 
Figure~\ref{fig:convergence_plots} demonstrates the test-time performance as a function of the algorithm's iterations (or layers in the unfolded networks). The unregularized deep models (\textit{w/o Reg}) display elevated variance and instability during the intermediate layer, as only the final performance is measured. However, with added regularization, the deep unfolded models remain stable and vastly accelerate convergence compared to the traditional Convex Bounding baselines. This is due to the large correction terms at the early iterations, as shown in Appendix \ref{sec:correction_analysis}.

\subsection{Robustness to Noise and Sparsity}
\label{sec:variable_params}

\begin{figure}[t]
    \centering
    \begin{subfigure}{\linewidth}
        \centering
        \includegraphics[width=\linewidth]{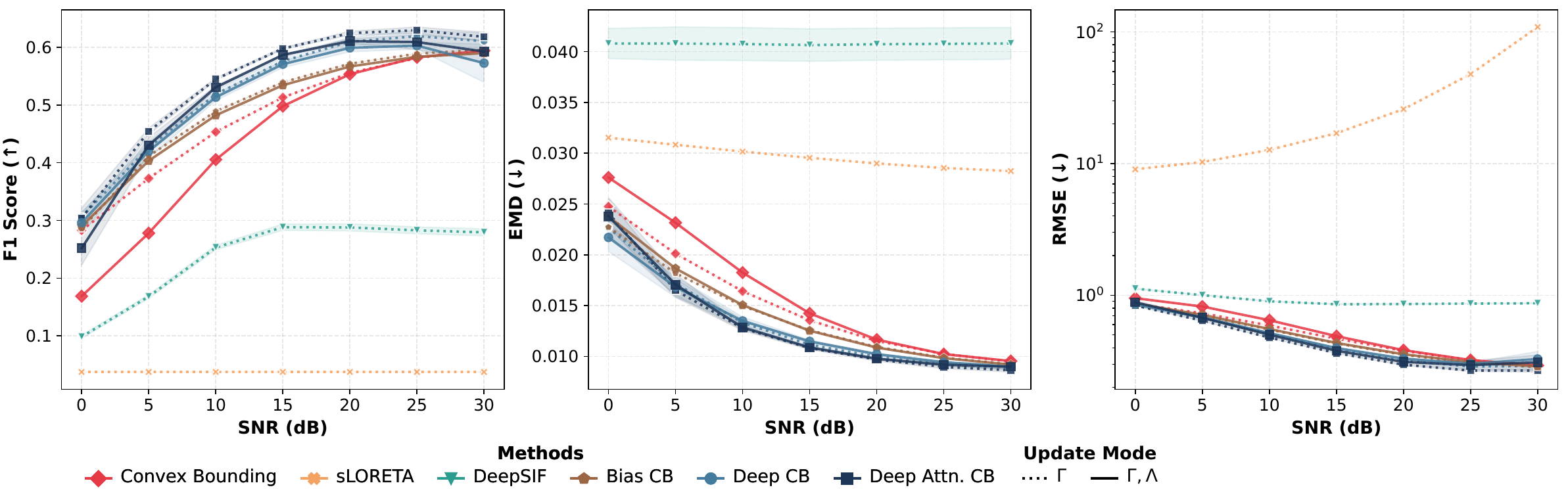}
        \vspace{-1em}
        \label{fig:snr_sweep}
    \end{subfigure}
    \vspace{0.5em}
    \begin{subfigure}{\linewidth}
        \centering
        \includegraphics[width=\linewidth]{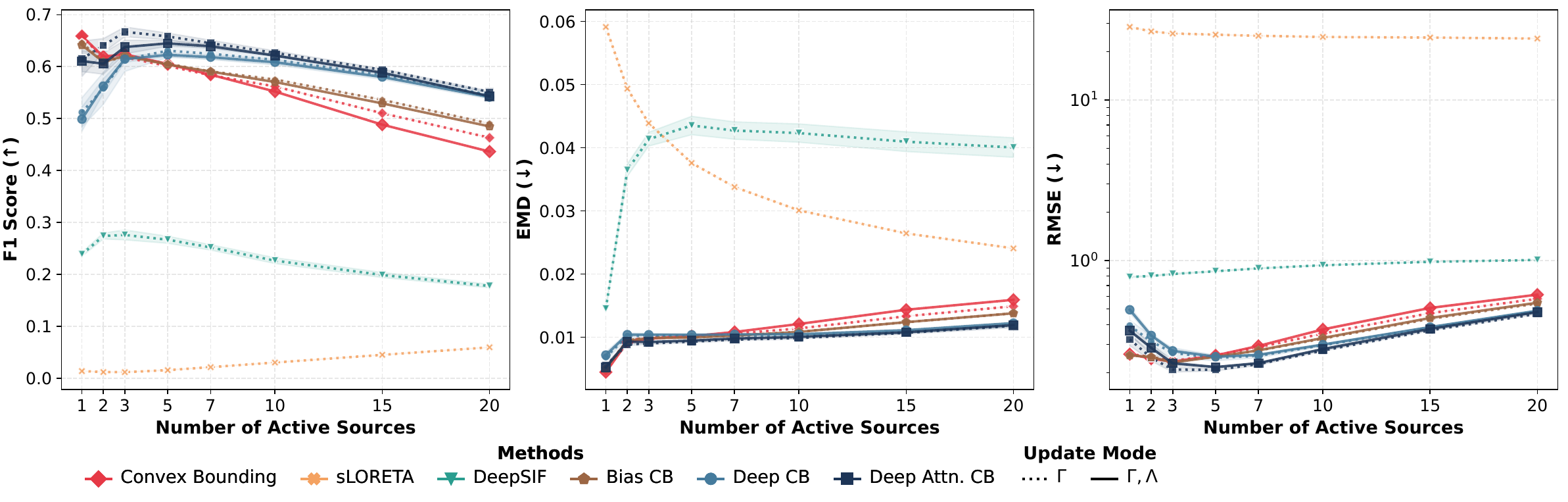}
        \vspace{-1em}
        \label{fig:nactive_sweep}
    \end{subfigure}
     \vspace{-2em}
    \caption{Robustness analysis. \textbf{Top:} Joint learning significantly improves performance at low SNRs. \textbf{Bottom:} Models show stable superiority within and beyond the training distribution ($5$--$20$ sources), with a minor data-driven bias dip for extremely sparse setups ($1$--$3$).}
    \label{fig:data_params}
    \vspace{-1em} 
\end{figure}

We evaluate model robustness against varying Signal-to-Noise Ratios (SNR) and active source counts (Figure~\ref{fig:data_params}). Across all noise levels, even low SNR out of the training distribution that was $\mathcal{U}(5,30)$, our correction learning approaches consistently outperform baselines. Crucially, the joint learning variants ($\boldsymbol{\Gamma}, \boldsymbol{\Lambda}$) match the performance of the $\boldsymbol{\Gamma}$-only variant that used the ground truth $\boldsymbol{\Lambda}$. 

When varying the number of active sources, we observe a slight decrease in performance for our deep methods in extremely sparse out-of-distribution settings ($1$--$3$ sources). This reflects a learned prior, as the models are biased toward the $5$--$20$ source range seen during training. Nonetheless, across all higher source counts, the unrolled models maintain highly stable and superior localization accuracy. We additionally compare the effect of source extent in Appendix \ref{sec:source_extent}.

\section{Limitations}
\label{sec:discussion}
First, our framework currently assumes fixed-orientation EEG sources and diagonal covariance models for both sources and sensors. While classical Type-II methods have explored richer noise structures, our unfolded architecture specifically targets the diagonal case to leverage its explicit scalar update rules. Extending correction learning to free-orientation dipoles, MEG-specific orientation models, and full-structure noise and source covariance matrices remains an area for future work.

\rebuttal{Second, global attention is more expressive and is strongest in several in-distribution settings, but its relative performance is less consistent under geometry and noise shifts, demonstrating its adaptivity by training across diverse lead fields remains future work. The subject-to-template experiment further shows that inaccurate anatomy remains a major failure mode (Appendix~\ref{app:head_model_generalization}). While learned corrections retain the Bayesian posterior computation and alternating positive-variance update structure, they are not guaranteed to retain the classical monotonic-descent property and should be viewed as finite-depth reconstruction networks anchored to CB (Appendix~\ref{app:typeii_objective}).}

\rebuttal{Third, reconstruction-accuracy evaluation still fundamentally relies on synthetic data and the learned modules retain biases tied to its sparsity, temporal, and head-model distributions. Our empirical THINGS-EEG2 study adds quantitative cross-session vertex and ROI consistency (Appendix~\ref{app:real_data_consistency}), but consistency is a repeatability proxy rather than direct localization accuracy. Without in-vivo cortical ground truth, such accuracy on real data cannot be established.}

Finally, the proposed correction-learning strategy has so far been evaluated solely on EEG source imaging. However, the underlying Type-II and sparse Bayesian learning mechanics are prevalent across sparse regression, classification, compressive sensing, and array-processing \cite{gerstoft2017sparse,ji2008bayesian,tipping2001sparse,wipf2007newview,wipf2004sparse}. Evaluating the transferability of our structure-preserving correction modules to these broader domains represents an interesting direction for future research.

\vspace{-0.8em}
\section{Conclusion}
We introduced a \emph{structure-preserving correction-learning framework} for sparse Type-II Bayesian source imaging. The proposed method unfolds the classical convex-bounding solver for source-prior and sensor-noise hyperparameter learning, represents its updates in a \textit{stable log-domain} form, and initializes the network so that it recovers the original solver before training. \textbf{Learned bias, residual MLP, or attention-augmented residual corrections}, then refine the hyperparameter-update dynamics without replacing the underlying Bayesian inference mechanism.

Across synthetic EEG benchmarks with heterogeneous sensor noise, our learned correction variants improve reconstruction quality, localization accuracy, support recovery, and convergence behavior compared to the \textit{classical convex-bounding baseline}. The strongest results are obtained by input-adaptive correction modules, especially Deep CB and Deep Attn. CB, while the joint source-noise variants retain performance close to the source-only setting with known noise. Furthermore, as detailed in our supplementary correction-term analysis, the learned modules mainly provide transient early-layer refinements and tend back toward the classical update dynamics in later layers. These results support \textit{residual correction learning} as a promising way to combine the \textit{interpretability of Type-II Bayesian solvers} with the \textit{empirical flexibility of trainable unfolded architectures}.

\bibliographystyle{plainnat}
\bibliography{references_R1}

\appendix

\section{Additional Architectural and Training Details}
\label{app:details}

\subsection{Stochastic deep supervision as training regularization}
\label{app:deep_supervision}

To stabilize optimization across many unfolded layers, we use \emph{stochastic deep supervision} as an auxiliary training regularizer \cite{lee2015deeply}. The main idea is to supplement the final reconstruction loss with additional losses on a small number of randomly selected intermediate layers. This helps reduce intermediate-state drift and improves gradient flow through the unfolded network, without incurring the cost of supervising all layers at every training step.

For unfolded layer $k=1,\dots,K$, let
\begin{equation}
\mathcal{J}_k(\Theta)
=
\frac{1}{T}\sum\limits_{t=1}^T
\left\|
\bar{\mathbf{x}}_\Theta^{k}(t)-\mathbf{x}^{\star}(t)
\right\|_2^2
\end{equation}
denote the reconstruction loss associated with the posterior mean at layer $k$. For each mini-batch, we sample uniformly without replacement a subset
\[
\mathcal{S}\subset\{1,\dots,K-1\}, \qquad |\mathcal{S}|=5,
\]
that is, five intermediate unfolded layers are chosen randomly from the available layers, excluding the final layer. The total training objective is then
\begin{equation}
\mathcal{L}_{\mathrm{total}}(\Theta)
=
\mathcal{J}_K(\Theta)
+
\beta_{\mathrm{DS}}\,
\frac{1}{|\mathcal{S}|}
\sum\limits_{k\in\mathcal{S}} \mathcal{J}_k(\Theta),
\label{eq:deep_supervision}
\end{equation}
where $\beta_{\mathrm{DS}}>0$ is the weight of the deep-supervision regularizer. In our experiments, selecting five random intermediate layers provided a good trade-off between computational overhead and diversity of auxiliary supervision signals.

\subsection{Bias-only and residual MLP correction modules}
\label{app:bias_mlp}

For completeness, we summarize the two local correction modules used in the main text. At unfolded layer $k$, recall the stabilized \emph{source-side} and \emph{noise-side} feature vectors
\begin{equation*}
\mathbf{h}_{\gamma,n}^{k}
=
\bigl[
u_n^{k},\,v_n^{k},\,\log(\gamma_n^{k}+\varepsilon_\gamma)
\bigr]^\top
\in \mathbb{R}^{3},
\qquad
\mathbf{h}_{\lambda,m}^{k}
=
\bigl[
r_m^{k},\,s_m^{k},\,\log(\lambda_m^{k}+\varepsilon_\lambda)
\bigr]^\top
\in \mathbb{R}^{3}.
\end{equation*}
These features define the local states from which the source-side and noise-side correction terms are predicted.

In the \emph{bias-only} variant, the correction is purely additive and input-independent:
\begin{equation}
c_{\gamma,n}^{k}=b_{\gamma,n}^{k}, \qquad
c_{\lambda,m}^{k}=b_{\lambda,m}^{k},
\end{equation}
where $\mathbf{b}_\gamma^{k}\in\mathbb{R}^{N}$ and $\mathbf{b}_\lambda^{k}\in\mathbb{R}^{M}$ are trainable bias vectors associated with unfolded layer $k$.

In the \emph{residual MLP} variant, the correction becomes input-dependent:
\begin{equation}
c_{\gamma,n}^{k}=f_\gamma^\theta(\mathbf{h}_{\gamma,n}^{k}), \qquad
c_{\lambda,m}^{k}=f_\lambda^\phi(\mathbf{h}_{\lambda,m}^{k}),
\end{equation}
where $f_\gamma^\theta:\mathbb{R}^{3}\to\mathbb{R}$ and $f_\lambda^\phi:\mathbb{R}^{3}\to\mathbb{R}$ are small pointwise multilayer perceptrons. The corresponding residual updates are
\begin{equation}
\ell_{\gamma,n}^{k+1}
=
\tfrac{1}{2}u_n^{k}-\tfrac{1}{2}v_n^{k}+c_{\gamma,n}^{k},
\qquad
\ell_{\lambda,m}^{k+1}
=
\tfrac{1}{2}r_m^{k}-\tfrac{1}{2}s_m^{k}+c_{\lambda,m}^{k},
\end{equation}
followed by $\gamma_n^{k+1}=\exp(\ell_{\gamma,n}^{k+1})$ and $\lambda_m^{k+1}=\exp(\ell_{\lambda,m}^{k+1})$.

In the source-only $\gamma$-approach, only the source-side correction module is used and the noise model remains fixed. In the joint $\gamma$-$\lambda$ approach, both correction modules are active and the two hyperparameter blocks are learned together within the alternating unfolded architecture.

To recover the classical convex-bounding baseline exactly at initialization, all correction modules are initialized to output zero. In the bias-only case, this means $b_{\gamma,n}^{k}=0$ and $b_{\lambda,m}^{k}=0$ for all $n,m,k$. In the MLP case, the final affine output layer is initialized to zero, so that the residual term vanishes initially and the unfolded network coincides exactly with the classical baseline.

\subsection{Attention-augmented source correction and optional sparsification}
\label{app:attention}

The classical convex-bounding source update is diagonal across sources, so each source variance is updated only from its own local statistics. To incorporate cross-source dependencies, we replace the pointwise source correction by an attention-augmented residual module. At unfolded layer $k$, for each currently considered source $n$ (typically the active subset), we define the local feature vector
\begin{equation}
\mathbf{h}_{\gamma,n}^{k}
=
\bigl[
u_n^{k},\,v_n^{k},\,\log(\gamma_n^{k}+\varepsilon_\gamma)
\bigr]^\top
\in\mathbb{R}^{3}.
\end{equation}
Stacking these row-wise yields the source-state matrix
\begin{equation}
\mathbf{H}_\gamma^{k}
=
\begin{bmatrix}
(\mathbf{h}_{\gamma,1}^{k})^\top\\
\vdots\\
(\mathbf{h}_{\gamma,N_{\mathrm{act}}}^{k})^\top
\end{bmatrix}
\in\mathbb{R}^{N_{\mathrm{act}}\times 3},
\end{equation}
where $N_{\mathrm{act}}$ denotes the number of sources processed by the \textit{attention block}.

The attention module consists of three stages: feature embedding, self-attention, and row-wise decoding. First, the input is projected into a $d$-dimensional latent space and normalized:
\begin{equation}
\mathbf{Z}^{(0)}
=
\operatorname{LayerNorm}\!\bigl(
\mathbf{H}_\gamma^{k}\mathbf{W}_{\mathrm{emb}}
+
\mathbf{1}\mathbf{b}_{\mathrm{emb}}^\top
\bigr),
\end{equation}
where $\mathbf{W}_{\mathrm{emb}}\in\mathbb{R}^{3\times d}$ and $\mathbf{b}_{\mathrm{emb}}\in\mathbb{R}^{d}$. Second, to capture cross-source interactions, we apply a $2$-head multi-head self-attention block with a residual skip connection:
\begin{equation}
\mathbf{Z}^{(1)}
=
\operatorname{LayerNorm}\!\Bigl(
\mathbf{Z}^{(0)}
+
\operatorname{MHSA}\bigl(\mathbf{Z}^{(0)},\mathbf{Z}^{(0)},\mathbf{Z}^{(0)}\bigr)
\Bigr).
\end{equation}
Finally, each contextualized row of $\mathbf{Z}^{(1)}$ is mapped to a scalar correction by a row-wise MLP with one hidden layer of width $2d$ and GELU activation:
\begin{equation}
\mathbf{c}_\gamma^{k}
=
g_\gamma^\theta(\mathbf{Z}^{(1)}),
\qquad
g_\gamma^\theta(\mathbf{z})
=
\mathbf{w}_{\mathrm{out}}^\top
\operatorname{GELU}\!\bigl(
\mathbf{W}_{\mathrm{hid}}\mathbf{z}
+
\mathbf{b}_{\mathrm{hid}}
\bigr)
+
b_{\mathrm{out}},
\end{equation}
where $\mathbf{W}_{\mathrm{hid}}\in\mathbb{R}^{2d\times d}$ and $\mathbf{w}_{\mathrm{out}}\in\mathbb{R}^{2d}$. The resulting source update retains the classical convex-bounding step as an explicit residual baseline:
\begin{equation}
\boldsymbol{\ell}_\gamma^{k+1}
=
\tfrac{1}{2}\mathbf{u}^{k}
-
\tfrac{1}{2}\mathbf{v}^{k}
+
\mathbf{c}_\gamma^{k},
\qquad
\boldsymbol{\gamma}^{k+1}
=
\exp(\boldsymbol{\ell}_\gamma^{k+1}).
\label{eq:attn_appendix}
\end{equation}
Thus, the attention module does not predict source updates from scratch; rather, it learns spatially informed correction terms around the analytical convex-bounding trajectory. In our experiments, the sensor-noise branch remains pointwise and is modeled by the residual MLP correction from Section~\ref{app:bias_mlp}. Multi-head self-attention is used here in the standard Transformer sense \cite{vaswani2017attention}.

To preserve exact recovery of the classical baseline at initialization, the final affine output layer of the attention-based correction module is initialized to zero, so that $\mathbf{c}_\gamma^{k}=\mathbf{0}$ before training.

When exact sparsity is desired, an optional hard-thresholding step may be applied after the source update. Given a relative threshold $\tilde{\tau}>0$, define
\begin{equation}
\tau^{k+1}
=
\tilde{\tau}\,
\max\limits_{1\leq n\leq N_{\mathrm{act}}}
\gamma_n^{k+1},
\end{equation}
and set
\begin{equation}
\gamma_n^{k+1}
\leftarrow
\begin{cases}
0, & \gamma_n^{k+1}<\tau^{k+1},\\
\gamma_n^{k+1}, & \text{otherwise}.
\end{cases}
\end{equation}
This step serves as an implementation-level sparsification mechanism; the core unfolded design remains the correction-based residual refinement of the classical convex-bounding update.

\subsection{Progressive layer-wise training and joint fine-tuning}
\label{app:progressive_training}

\begin{figure}[!t]
    \centering
    \includegraphics[width=0.8\linewidth]{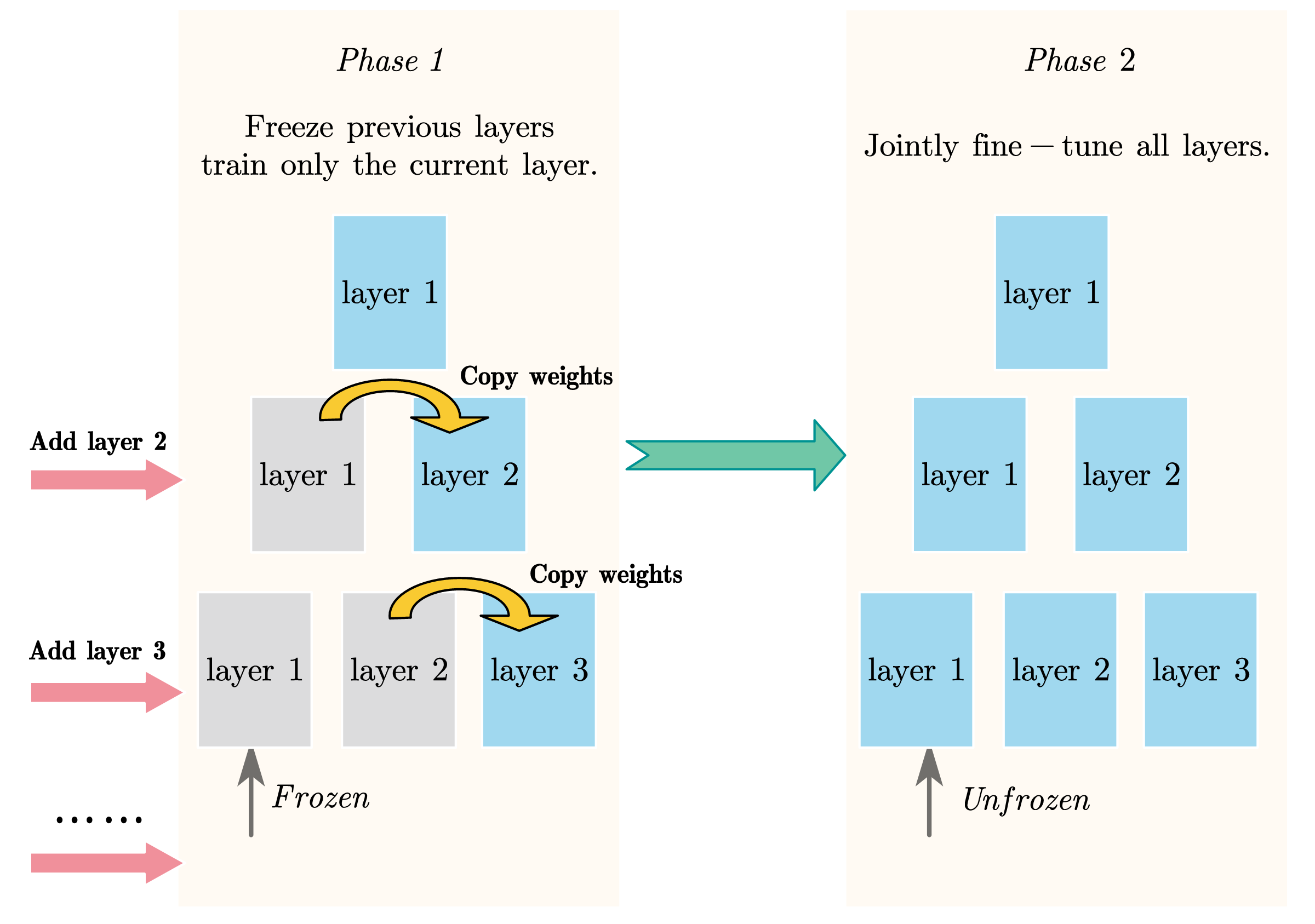}
    \caption{Two-phase training strategy for the unfolded architecture. After adding a new layer, previously trained layers are frozen and only the newly added layer is optimized, with initialization copied from the preceding trained layer. After 2 epochs, all layers are unfrozen and the unfolded network is jointly fine-tuned end-to-end for another 2 epochs, before adding a new layer.}
    \label{fig:progressive_training}
\end{figure}

To improve optimization stability for deeper unfolded architectures, we use a two-phase training strategy illustrated in Figure~\ref{fig:progressive_training}. Rather than training the full $K$-layer unfolded network from scratch, we first increase the model depth progressively. When unfolded layer $k$ is added, all previously trained layers $1,\dots,k-1$ are frozen and only the current layer is trained. To obtain a stable initialization, the parameters of the newly added layer are copied from the most recently trained preceding layer. Afterwards, all $1,\dots,k$ layers are unfrozen and the full network is jointly fine-tuned end-to-end using the loss from the main text, together with the stochastic deep-supervision regularizer from Appendix~\ref{app:deep_supervision}. This schedule is particularly useful for the deeper correction-learning variants, where training all layers simultaneously from the beginning can lead to unstable optimization.

\section{Experimental Setup Details}
\label{app:exp_setup}

\subsection{Synthetic source and noise generation}
\label{app:synthetic_generation}

We generate synthetic EEG measurements by combining sparse spatiotemporal source activity with a realistic forward model. Following the general simulation philosophy of \cite{morik2024physics3d}, each sample contains a randomly chosen number of active sources with focal spatial support. The spatial footprint of each source is obtained from a Gaussian kernel on cortical Euclidean distances, using a small standard deviation to encourage spatial sparsity and focal activations. 
To emulate realistic neural oscillations, the temporal activation for each active source is synthesized by randomly assigning it to one of five canonical frequency bands ($\delta, \theta, \alpha, \mu$, or $\beta$). We generate independent white Gaussian noise of length $T$ and apply a bandpass filter corresponding to the selected band. The full source matrix $\mathbf{X}$ is finally obtained by superposing these spatiotemporal components.

The sensor data are generated through
\[
\mathbf{Y}=\mathbf{L}\mathbf{X}+\mathbf{E}.
\]
To test the value of joint adaptive noise learning, we corrupt the sensor measurements with diagonal heteroscedastic Gaussian noise. Specifically, each sensor is assigned an independent SNR sampled uniformly from a prescribed interval, which determines the corresponding diagonal noise variance in $\boldsymbol{\Lambda}$. This yields a controlled benchmark for comparing source-only and joint source-noise inference under sensor-wise heterogeneous noise.
We summarize our hyperparameters used in this work in Table \ref{tab:sim_params}.

\begin{table}[t]
\centering
\caption{Hyperparameters for synthetic M/EEG data generation.}
\label{tab:sim_params}
\begin{tabular}{ll}
\toprule
\textbf{Parameter Category} & \textbf{Value / Distribution} \\
\midrule
\multicolumn{2}{l}{\textbf{Dataset \& Head Model}} \\
Head Model & \texttt{fsaverage} \\
Source Space & Surface (\texttt{ico3}), Fixed Orientation \\
Training Samples & 100,000 \\
Validation Samples & 1,000 per evaluation split \\
\midrule
\multicolumn{2}{l}{\textbf{Temporal Dynamics}} \\
Sampling Frequency ($f_s$) & 100 Hz \\
Time Steps ($T$) & 100 \\
Delta band & 0.5--4 Hz \\
Theta band & 4--8 Hz \\
Alpha band & 8--13 Hz \\
Mu band & 8--12 Hz \\
Beta band & 12--30 Hz \\
\midrule
\multicolumn{2}{l}{\textbf{Spatial \& Source Parameters}} \\
Number of Active Sources & $\mathcal{U}(5,20)$ \\
Spatial Spread (Std. Dev.) & $\mathcal{U}(0.001,0.005)$ \\
\midrule
\multicolumn{2}{l}{\textbf{Noise Characteristics}} \\
Channel SNR Range & $\mathcal{U}(5,30)$ dB \\
Heteroscedasticity & Independent SNR sampled per sensor \\
\bottomrule
\end{tabular}
\end{table}

\subsection{Forward model and inverse-crime mitigation}
\label{app:forward_model}

We construct the forward model on the \texttt{fsaverage} template brain \cite{fischl1999high} using MNE-Python \cite{GramfortEtAl2013a}. The cortical source space is discretized on an \texttt{ico3} mesh ($1284$ sources) with fixed dipole orientations constrained normal to the cortical surface. The lead-field matrix $\mathbf{L}$ is computed using a three-layer boundary element model (BEM) \cite{fuchs2002standardized}. To mitigate the inverse crime \cite{colton1998inverse,kaipio2007statistical}, validation data are generated under controlled forward-model mismatch by perturbing tissue conductivities, following established practice in EEG source-imaging benchmarks \cite{hecker2021convdip,wolters2010combined}. Unless otherwise specified, all core experiments throughout the paper are evaluated under this standard configuration, which we denote as the \textbf{Train Cap, Ico3} setting.

To evaluate out-of-distribution generalization, we introduce two additional transfer settings based on the THINGS-EEG sensor layout. These settings are designed to probe algorithmic robustness to unseen sensor layouts, longer temporal contexts, and increased spatial resolutions:

\begin{itemize}
    \item \textbf{Things Cap, Ico3}: A new lead field is generated for the THINGS sensor configuration, introducing an explicit forward-model mismatch relative to the synthetic training setup. Additionally, the temporal window is extended to be six times longer ($600$ timesteps instead of $100$).
    \item \textbf{Things Cap, Ico4}: This setting builds upon the previous one by additionally increasing the source-space granularity at evaluation time. Models trained on the $1284$ sources of the \texttt{ico3} mesh must perform zero-shot inference on an \texttt{ico4} mesh containing $5124$ sources.
\end{itemize}

\subsection{Evaluation metrics}
\label{app:metrics}

We evaluate reconstruction quality along three complementary axes. 
For signal fidelity, we report relative mean squared error (rMSE),
\[
\mathrm{rMSE}
=
\frac{\|\hat{\mathbf{X}}-\mathbf{X}\|_F^2}{\|\mathbf{X}\|_F^2}.
\]

For spatial localization, we report localization error (LE) and Earth Mover's Distance (EMD) that rely on the 3D source coordinates $\mathbf{r}$. LE measures for every true local maximum $i \in \mathcal{M}_{\text{true}}$ the Euclidean distance to the nearest predicted maximum $j \in \mathcal{M}_{\text{pred}}$:$$
\mathrm{LE} = \frac{1}{|\mathcal{M}_{\text{true}}|} \sum_{i \in \mathcal{M}_{\text{true}}} \min_{j \in \mathcal{M}_{\text{pred}}} \|\mathbf{r}_i - \mathbf{r}_j\|_2.
$$
To capture the overall spatial dispersion rather than just peak-to-peak distances, we compute the Earth Mover's Distance (EMD). Let $m_i$ and $\hat{m}_j$ represent the normalized temporal $L_2$-norms (masses) of the active true and predicted sources, such that $\sum_{i} m_i = 1$ and $\sum_{j} \hat{m}_j = 1$. The EMD computes the optimal transport cost between these spatial distributions:
$$
\mathrm{EMD} = \min_{\mathbf{T} \ge 0} \sum_{i \in \mathcal{A}_{\text{true}}} \sum_{j \in \mathcal{A}_{\text{pred}}} T_{i,j} \|\mathbf{r}_i - \mathbf{r}_j\|_2
$$
subject to the marginal constraints $\sum_j T_{i,j} = m_i$ and $\sum_i T_{i,j} = \hat{m}_j$.

For support recovery, we report the F1-score after thresholding temporal root-mean-square source amplitudes to determine active dipoles:
$$
\mathrm{F1} = \frac{2 \cdot \text{TP}}{2 \cdot \text{TP} + \text{FP} + \text{FN}}.
$$

\subsection{Baselines and implementation details}
\label{app:baselines}

We compare the proposed unfolded correction-learning models against four baseline families. First, we include sLORETA as a standard linear minimum-norm reference \cite{pascual2002standardized}. Second, we include classical source-only $\gamma$-MAP, which performs
sparse Type-II inference while using the known diagonal noise covariance $\boldsymbol{\Lambda}_0$ \cite{wipf2009unified}. Third, we include the classical joint $\gamma$-$\lambda$ convex-bounding solver with adaptive diagonal noise learning \cite{cai2021robust}, which serves as the non-learned model-based baseline for our joint unfolding approach.
\rebuttal{We also compare to two different Type-II optimization algorithms that have different update rules \cite{cai2021robust}:
$$ \mathrm{EM}: \quad \gamma_n^{(k+1)}=e_n^k+[\boldsymbol{\Sigma}_\mathbf{x}^k]_{nn}, \qquad \lambda_m^{(k+1)}=d_m^k+\lambda_m^k-(\lambda_m^k)^2q_m^k, $$
$$ \mathrm{MacKay}: \quad \displaystyle\gamma_n^{(k+1)}=\frac{e_n^k}{\gamma_n^kz_n^k}, \qquad \displaystyle\lambda_m^{(k+1)}=\frac{d_m^k}{\lambda_m^kq_m^k}, $$
We ran these classical algorithms for 250 iterations to ensure convergence. Next, we include a noise-whitened Group Lasso, also known as the mixed-norm estimate (MxNE)\cite{gramfort2012mixed}:
$$ \widehat{\mathbf X}=\arg\min_{\mathbf X}\frac{1}{2}\left\|\boldsymbol{\Lambda}_0^{-1/2}(\mathbf L\mathbf X-\mathbf Y)\right\|_F^2+\alpha\sum_{n=1}^{N}\left\|\mathbf X_{n,:}\right\|_2. $$
We solve this convex objective using FISTA. Its $\ell_{2,1}$ penalty promotes source-wise sparsity by jointly selecting complete source time courses. We set $\alpha=0.001\alpha_{\max}$, where $\alpha_{\max}=\max_n\left\|\mathbf l_n^\top\boldsymbol{\Lambda}_0^{-1}\mathbf Y\right\|_2$ with $\mathbf l_n$ denoting the $n$th column of the lead-field
matrix $\mathbf L$, after tuning the ratio $\alpha/\alpha_{\max}$ on 100 synthetic training examples by minimizing the mean relative source MSE.
}
Finally, we compare against DeepSIF as a representative end-to-end deep source-imaging model \cite{sun2022deepsif}.

We list the training hyperparameters in Table \ref{tab:train_params}. Further implementation details and training schedules are given with the supplementary code.
\begin{table}[t]
\centering
\caption{Hyperparameters for model architecture and training.}
\label{tab:train_params}
\begin{tabular}{ll}
\toprule
\textbf{Hyperparameter} & \textbf{Value} \\
\midrule
\multicolumn{2}{l}{\textbf{Architecture}} \\
Unrolled Layers ($K$) & 25 \\
Embedding Dimension ($d$) & 16 \\
Attention Heads & 2 \\
MLP Hidden Dimension & 32 ($2 \times d$) \\
\midrule
\multicolumn{2}{l}{\textbf{Optimization}} \\
Optimizer & Adam\\
Learning Rate & $0.001$ \\
Adam $\beta_1, \beta_2$ & $0.9, 0.999$ \\
Batch Size & 512 \\
Epochs per Training Phase & 2 \\
Total Epochs & 100 \\
Gradient Norm Clipping & 1 \\
Relative Sparsity Threshold ($\tilde{\tau}$) & $10^{-4}$ \\
Deep Supervision Weight ($\beta_{\text{DS}}$) & $0.5$ \\
Deep Supervision Layers & Max 5 (Stochastically Sampled) \\
\bottomrule
\end{tabular}
\end{table}

\subsection{Computational Resources Used}
\label{appx:computeresources}
All experiments were executed on a small compute cluster, with each individual training job utilizing a single NVIDIA A100 40GB GPU. Training a single model for 100 epochs took approximately 2.5 hours for Bias CB ($\boldsymbol{\Gamma}$) and 4 hours for Bias CB ($\boldsymbol{\Gamma},\boldsymbol{\Lambda}$); 3 hours for Deep CB ($\boldsymbol{\Gamma}$) and 4.5 hours for Deep CB ($\boldsymbol{\Gamma},\boldsymbol{\Lambda}$); and 14 hours for Deep Attn. CB ($\boldsymbol{\Gamma}$) and 15 hours for Deep Attn. CB ($\boldsymbol{\Gamma},\boldsymbol{\Lambda}$). Training the end-to-end DeepSIF baseline took 8 hours until convergence.

To limit the memory footprint of our deep unrolled architecture (25 layers), we utilized PyTorch gradient checkpointing at the cost of additional training time. In contrast, inference is highly efficient, requiring less than 0.1 seconds per sample on the test set. The total estimated compute time for the entire project (including method development, preliminary hyperparameter tuning, and final results) was approximately 2,000 GPU hours, while only recreating the results reported in the paper, including the multi-seed evaluations reported in Table \ref{tab:results_final}, takes around 400 GPU hours.

\section{Additional Results}

\subsection{Comprehensive Performance Overview}
\label{sec:general_performance}

Table~\ref{tab:results_final} summarizes the robustness and generalization performance of our proposed methods under increasingly challenging out-of-distribution (OOD) settings.

Single-step approaches such as sLORETA and end-to-end deep models such as DeepSIF struggle to accurately recover sparse sources. Moreover, standard deep architectures are structurally tied to fixed sensor and source dimensions, limiting their applicability to unseen configurations. This dimensional rigidity also affects our simplified Biased CB variants: although both Biased CB and Biased CB (joint) outperform the standard Convex Bounding baseline, they fail to generalize when the source space resolution changes.

In contrast, our unfolded Deep CB and Deep Attn. CB architectures overcome these limitations. By learning an input-dependent correction term with weights shared across sensors or sources, these models naturally generalize to unseen sensor layouts and source resolutions. Across all evaluation settings, the unfolded architectures consistently achieve the strongest overall performance, with Deep Attn. CB generally providing the best results. 

The joint learning variants $(\boldsymbol{\Gamma},\boldsymbol{\Lambda})$ further demonstrate that dynamically estimating the noise covariance introduces only a marginal performance degradation compared to using the ground-truth noise covariance when predicting only $\boldsymbol{\Gamma}$, highlighting the robustness of the joint optimization procedure.
\rebuttal{Comparing Type-II optimizers, EM performs worse, while MacKay achieved slightly better EMD on the Things Caps.}

Although iterative deep unrolling incurs a higher computational cost than single-step feedforward inference, both Deep CB and Deep Attn. CB remain computationally efficient, requiring well below 0.1 seconds per sample at inference time.

\begin{table}[t]
\centering
\caption{Performance comparison (mean $\pm$ std over 5 seeds for learned methods).}
\label{tab:results_final}
\small
\begin{tabular}{lccccc}
\toprule
Method & rMSE $\downarrow$ & EMD $\downarrow$ & F1 $\uparrow$ & LE $\downarrow$ & Time $\downarrow$ \\
\midrule\multicolumn{6}{l}{\textbf{Train Cap}} \\
\midrule
Convex Bounding ($\boldsymbol{\Gamma}$) & 0.410 & 0.0122 & 0.535 & 0.0056 & 0.0048 \\
Convex Bounding ($\boldsymbol{\Gamma}, \boldsymbol{\Lambda}$) & 0.434 & 0.0130 & 0.522 & 0.0060 & 0.0046 \\
MacKay ($\boldsymbol{\Gamma}$) & 0.413 & 0.0105 & 0.524 & 0.0051 & 0.0083 \\
MacKay ($\boldsymbol{\Gamma}, \boldsymbol{\Lambda}$) & 0.424 & 0.0106 & 0.512 & 0.0053 & 0.0182 \\
EM ($\boldsymbol{\Gamma}$) & 0.475 & 0.0163 & 0.524 & 0.0067 & 0.0085 \\
EM ($\boldsymbol{\Gamma}, \boldsymbol{\Lambda}$) & 0.491 & 0.0166 & 0.509 & 0.0069 & 0.0181 \\
Group Lasso & 0.946 & 0.0269 & 0.140 & 0.0166 & 0.0528 \\
sLORETA & 24.373 & 0.0291 & 0.038 & 0.0251 & 0.0010 \\
DeepSIF & 0.796 \tiny{$\pm$0.007} & 0.0403 \tiny{$\pm$0.0018} & 0.331 \tiny{$\pm$0.009} & 0.0104 \tiny{$\pm$0.0001} & \textbf{0.0003 \tiny{$\pm$0.0001}} \\
Bias CB ($\boldsymbol{\Gamma}$) & 0.383 \tiny{$\pm$0.000} & 0.0115 \tiny{$\pm$0.0000} & 0.553 \tiny{$\pm$0.000} & 0.0054 \tiny{$\pm$0.0000} & 0.0025 \tiny{$\pm$0.0007} \\
Bias CB ($\boldsymbol{\Gamma}, \boldsymbol{\Lambda}$) & 0.386 \tiny{$\pm$0.000} & 0.0115 \tiny{$\pm$0.0000} & 0.548 \tiny{$\pm$0.000} & 0.0054 \tiny{$\pm$0.0000} & 0.0047 \tiny{$\pm$0.0000} \\
Deep CB ($\boldsymbol{\Gamma}$) & 0.344 \tiny{$\pm$0.006} & 0.0105 \tiny{$\pm$0.0002} & 0.594 \tiny{$\pm$0.008} & 0.0048 \tiny{$\pm$0.0000} & 0.0027 \tiny{$\pm$0.0000} \\
Deep CB ($\boldsymbol{\Gamma}, \boldsymbol{\Lambda}$) & 0.348 \tiny{$\pm$0.005} & 0.0108 \tiny{$\pm$0.0001} & 0.588 \tiny{$\pm$0.006} & 0.0048 \tiny{$\pm$0.0001} & 0.0063 \tiny{$\pm$0.0023} \\
Deep Attn. CB ($\boldsymbol{\Gamma}$) & \textbf{0.323 \tiny{$\pm$0.002}} & \textbf{0.0102 \tiny{$\pm$0.0002}} & \textbf{0.608 \tiny{$\pm$0.003}} & \textbf{0.0045 \tiny{$\pm$0.0000}} & 0.0071 \tiny{$\pm$0.0001} \\
Deep Attn. CB ($\boldsymbol{\Gamma}, \boldsymbol{\Lambda}$) & 0.332 \tiny{$\pm$0.002} & \textbf{0.0103 \tiny{$\pm$0.0001}} & 0.600 \tiny{$\pm$0.001} & 0.0046 \tiny{$\pm$0.0000} & 0.0090 \tiny{$\pm$0.0001} \\
\midrule\multicolumn{6}{l}{\textbf{Things Cap Ico3}} \\
\midrule
Convex Bounding ($\boldsymbol{\Gamma}$) & 0.384 & 0.0127 & 0.722 & 0.0056 & 0.0030 \\
Convex Bounding ($\boldsymbol{\Gamma}, \boldsymbol{\Lambda}$) & 0.408 & 0.0135 & 0.703 & 0.0060 & 0.0062 \\
MacKay ($\boldsymbol{\Gamma}$) & 0.358 & \textbf{0.0101} & 0.738 & 0.0048 & 0.0119 \\
MacKay ($\boldsymbol{\Gamma}, \boldsymbol{\Lambda}$) & 0.366 & \textbf{0.0101} & 0.734 & 0.0048 & 0.0263 \\
EM ($\boldsymbol{\Gamma}$) & 0.476 & 0.0181 & 0.664 & 0.0070 & 0.0119 \\
EM ($\boldsymbol{\Gamma}, \boldsymbol{\Lambda}$) & 0.493 & 0.0185 & 0.651 & 0.0072 & 0.0264 \\
Group Lasso & 0.929 & 0.0267 & 0.224 & 0.0152 & 0.3321 \\
sLORETA & 64.609 & 0.0302 & 0.024 & 0.0239 & \textbf{0.0009} \\
Bias CB ($\boldsymbol{\Gamma}$) & 0.358 \tiny{$\pm$0.000} & {0.0121 \tiny{$\pm$0.0000}} & 0.742 \tiny{$\pm$0.000} & 0.0053 \tiny{$\pm$0.0000} & 0.0031 \tiny{$\pm$0.0007} \\
Deep CB ($\boldsymbol{\Gamma}$) & 0.556 \tiny{$\pm$0.144} & {0.0125 \tiny{$\pm$0.0005}} & 0.683 \tiny{$\pm$0.042} & 0.0067 \tiny{$\pm$0.0006} & 0.0022 \tiny{$\pm$0.0000} \\
Deep CB ($\boldsymbol{\Gamma}, \boldsymbol{\Lambda}$) & 0.393 \tiny{$\pm$0.008} & 0.0131 \tiny{$\pm$0.0002} & 0.745 \tiny{$\pm$0.006} & 0.0056 \tiny{$\pm$0.0002} & 0.0049 \tiny{$\pm$0.0014} \\
Deep Attn. CB ($\boldsymbol{\Gamma}$) & \textbf{0.334 \tiny{$\pm$0.009}} & {0.0117 \tiny{$\pm$0.0007}} & \textbf{0.765 \tiny{$\pm$0.006}} & \textbf{0.0048 \tiny{$\pm$0.0002}} & 0.0036 \tiny{$\pm$0.0001} \\
Deep Attn. CB ($\boldsymbol{\Gamma}, \boldsymbol{\Lambda}$) & 0.351 \tiny{$\pm$0.020} & {0.0120 \tiny{$\pm$0.0010}} & 0.754 \tiny{$\pm$0.012} & 0.0051 \tiny{$\pm$0.0002} & 0.0056 \tiny{$\pm$0.0000} \\
\midrule\multicolumn{6}{l}{\textbf{Things Cap Ico4}} \\
\midrule
Convex Bounding ($\boldsymbol{\Gamma}$) & 1.182 & 0.0170 & 0.286 & 0.0111 & 0.0264 \\
Convex Bounding ($\boldsymbol{\Gamma}, \boldsymbol{\Lambda}$) & 1.188 & 0.0177 & 0.270 & 0.0115 & 0.0442 \\
MacKay ($\boldsymbol{\Gamma}$) & 1.594 & \textbf{0.0133} & 0.291 & 0.0093 & 0.0341 \\
MacKay ($\boldsymbol{\Gamma}, \boldsymbol{\Lambda}$) & 1.592 & \textbf{0.0134} & 0.286 & 0.0094 & 0.0731 \\
EM ($\boldsymbol{\Gamma}$) & 1.215 & 0.0348 & 0.126 & 0.0141 & 0.0348 \\
EM ($\boldsymbol{\Gamma}, \boldsymbol{\Lambda}$) & 1.223 & 0.0344 & 0.123 & 0.0141 & 0.0749 \\
Group Lasso & 1.201 & 0.0270 & 0.066 & 0.0187 & 0.8830 \\
sLORETA & 83.035 & 0.0267 & 0.023 & 0.0265 & \textbf{0.0101} \\
Deep CB ($\boldsymbol{\Gamma}$) & \textbf{1.092 \tiny{$\pm$0.030}} & 0.0154 \tiny{$\pm$0.0006} & 0.287 \tiny{$\pm$0.004} & 0.0093 \tiny{$\pm$0.0002} & 0.0285 \tiny{$\pm$0.0001} \\
Deep CB ($\boldsymbol{\Gamma}, \boldsymbol{\Lambda}$) & \textbf{1.078 \tiny{$\pm$0.053}} & 0.0153 \tiny{$\pm$0.0006} & 0.282 \tiny{$\pm$0.014} & 0.0094 \tiny{$\pm$0.0003} & 0.0591 \tiny{$\pm$0.0066} \\
Deep Attn. CB ($\boldsymbol{\Gamma}$) & \textbf{1.122 \tiny{$\pm$0.066}} & 0.0148 \tiny{$\pm$0.0005} & \textbf{0.313 \tiny{$\pm$0.009}} & \textbf{0.0084 \tiny{$\pm$0.0002}} & 0.0714 \tiny{$\pm$0.0008} \\
Deep Attn. CB ($\boldsymbol{\Gamma}, \boldsymbol{\Lambda}$) & 1.184 \tiny{$\pm$0.075} & 0.0156 \tiny{$\pm$0.0009} & 0.296 \tiny{$\pm$0.007} & 0.0090 \tiny{$\pm$0.0000} & 0.0978 \tiny{$\pm$0.0002} \\
\midrule
\bottomrule
\bottomrule
\end{tabular}
\end{table}




\subsection{Importance of Source Extent}
\label{sec:source_extent}
\begin{figure}[t]
    \centering
    \includegraphics[width=\linewidth]{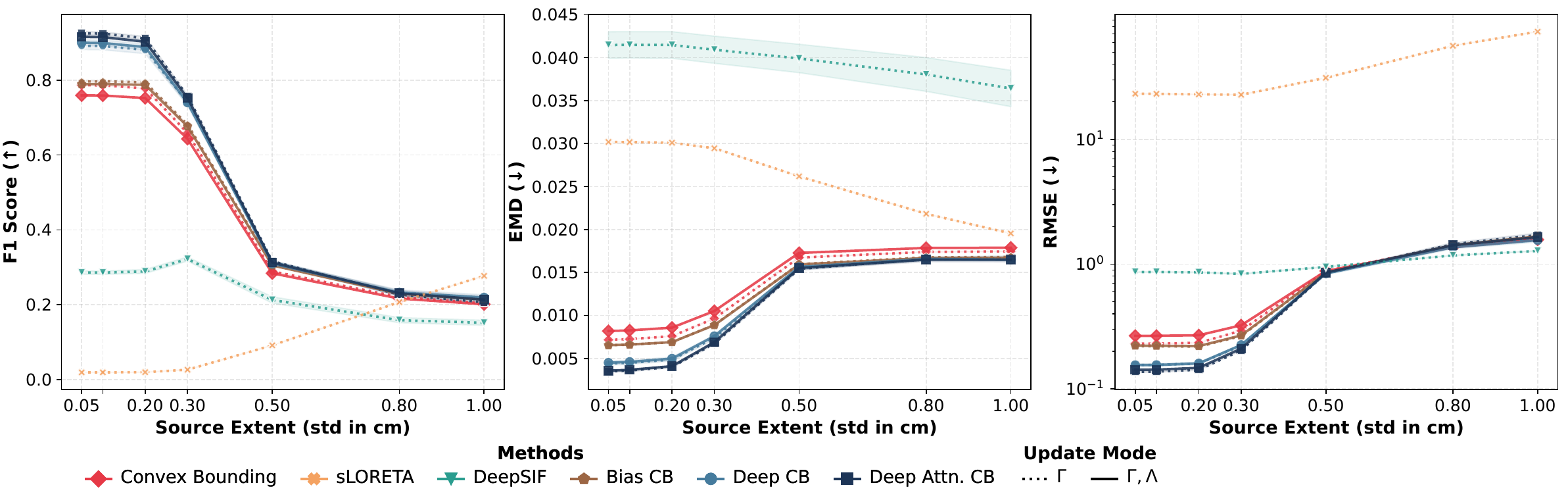}
    \caption{Performance as a function of the underlying ground-truth source extent (standard deviation in cm). While the SBL-based deep models dominate for focal sources and naturally drop in exact support recovery (F1) for diffuse activity, their spatial (EMD) and magnitude (rMSE) errors remain competitive, demonstrating robustness across spatial scales.}
    \label{fig:sourceextent}
\end{figure}

Our proposed architectures are grounded in the SBL framework; they inherently assume localizing highly focal source activity. Figure~\ref{fig:sourceextent} illustrates performance across varying simulated source extents (standard deviation in cm). For spatially restricted sources ($\le$ 0.30 cm), the deep unrolled models (Deep CB and Deep Attn. CB) achieve high F1 scores and minimal spatial error (EMD), while traditional dense methods like sLORETA/DeepSIF exhibit the opposite behavior. They struggle to accurately recover sparse sources but show improving F1 and EMD metrics as the active region becomes larger and more diffuse, reflecting their structural bias toward spatially smooth solutions. 

Crucially, our methods remain robust even when the underlying sparsity assumption is heavily relaxed. While exact support recovery (F1 score) naturally declines for widespread sources (> 0.50 cm), the spatial distribution metrics (EMD and rMSE) plateau and remain competitive with the baselines. This demonstrates that even for large source extents, the deep SBL architectures predict compact representations of the broader active cortical region.

\subsection{Correction Term Analysis}
\label{sec:correction_analysis}

\begin{figure}[t]
    \centering
    \includegraphics[width=\linewidth]{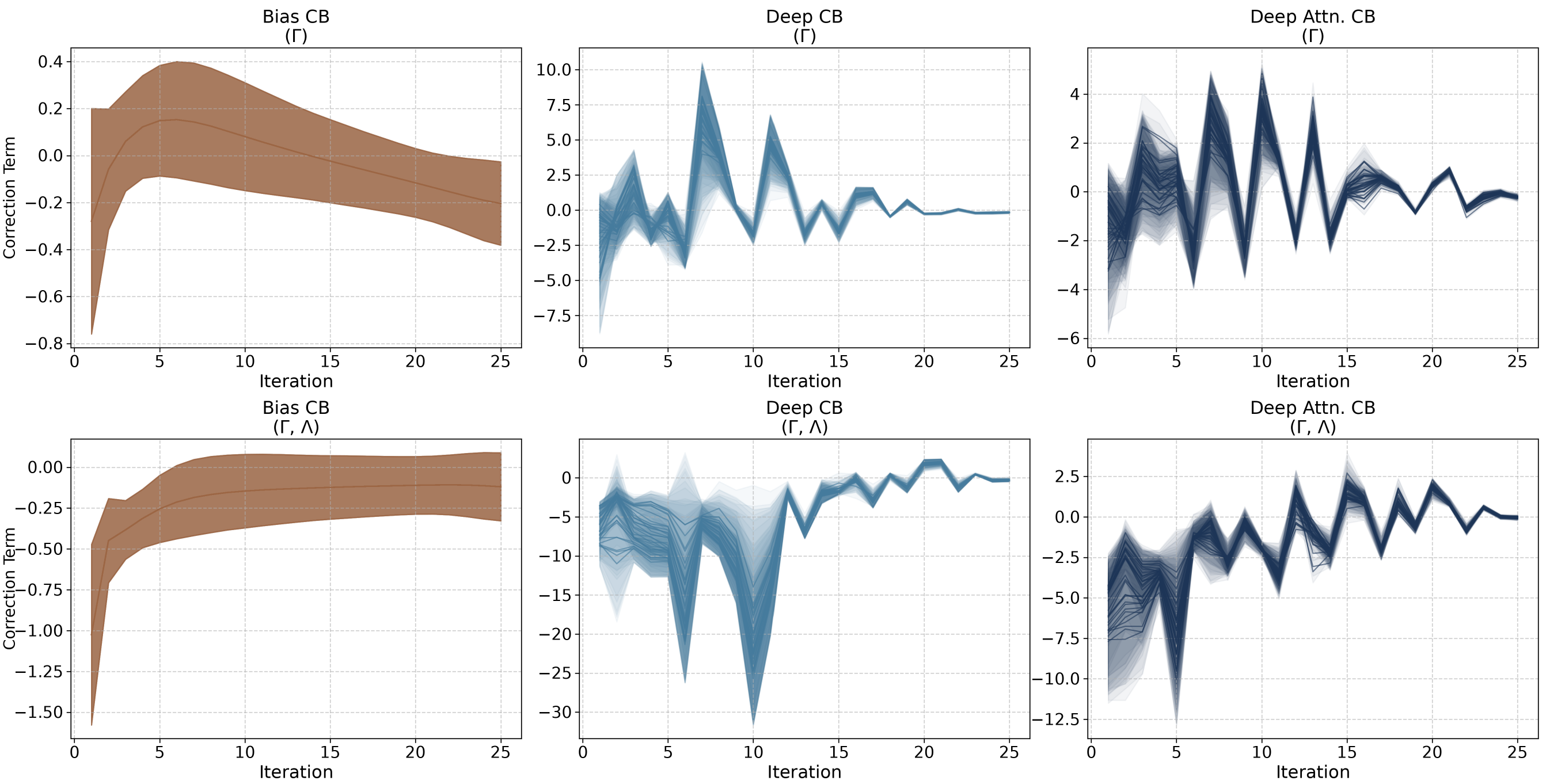}
    \caption{Evolution of \emph{learned correction terms} across algorithmic iterations for Bias CB, Deep CB, and Deep Attn. CB. Results are shown for 32 random test samples, with solid lines indicating sample-wise mean corrections and shaded regions indicating variability over sources. \textbf{Top:} Source covariance learning only $(\boldsymbol{\Gamma})$. \textbf{Bottom:} Joint source and noise learning $(\boldsymbol{\Gamma},\boldsymbol{\Lambda})$. The deep correction variants show larger early-layer corrections that shrink toward zero in later layers, indicating that the learned modules provide transient refinements while the final updates move closer to the classical convex-bounding dynamics.}
    \label{fig:correction_terms}
\end{figure}

The unfolded architecture makes it possible to inspect how the learned correction terms modify the classical convex-bounding update across layers. Figure~\ref{fig:correction_terms} visualizes the trajectory of these correction terms across iterations.

For Bias CB, the correction trajectories are relatively smooth because the corrections are input-independent. In contrast, Deep CB and Deep Attn. CB produce input-dependent corrections with substantially larger early-layer variability across sources and test samples. This behavior is most pronounced in the joint source-noise setting $(\boldsymbol{\Gamma},\boldsymbol{\Lambda})$, where the early corrections are often strongly negative before gradually shrinking in magnitude.

A key observation is that, for the deep correction variants, the correction terms tend toward zero over the 25 unfolded layers on the test set. Since the classical convex-bounding update is recovered when the learned correction is zero, this suggests that the neural modules mainly act as transient refinements of the hyperparameter trajectory rather than replacing the baseline Bayesian update throughout the full inference process. This interpretation is consistent with the quantitative results, where the best performance is obtained at the final unfolded iteration: as the reconstruction improves, the learned dynamics move closer to the original convex-bounding update after earlier corrective fluctuations.

Thus, the correction plots provide an interpretable link between the \textbf{baseline solver} and the \textbf{learned unfolded models}. The learned modules steer the early update dynamics, especially when both source and noise hyperparameters are adapted, while the later layers increasingly resemble the classical convex-bounding behavior.

\subsection{Application to Real-World Data}
\label{sec:realworlddata}
To demonstrate that our unfolded networks generalize beyond simulated conditions, we evaluate them zero-shot on empirical data from the THINGS-EEG2 dataset \cite{gifford2022large}. This dataset contains high-density EEG recordings of subjects viewing diverse object images, providing a rigorous testbed for localizing visual evoked potentials (VEPs). Evaluating M/EEG source localization on empirical data is notoriously challenging due to the strict absence of ground-truth source activity. Consequently, this validation assesses whether our models can recover established neurophysiological priors, specifically, that early visual stimuli should reliably evoke focal responses in the occipital lobe.
We extract the first 200ms post-stimulus window to capture the primary visual response. Because real-world EEG suffers from inherently low signal-to-noise ratios and severe sensor-wise noise fluctuations, we evaluate the inverse solvers under two regimes: \textit{trial-averaged} data (where background noise is suppressed) and strictly \textit{single-trial} data (a highly challenging, low-SNR regime). All network weights are frozen from the synthetic training phase; no real-world fine-tuning is performed.

\begin{figure}[ht]
    \centering
    
    \begin{subfigure}[b]{0.48\linewidth}
        \centering
        \includegraphics[width=\linewidth]{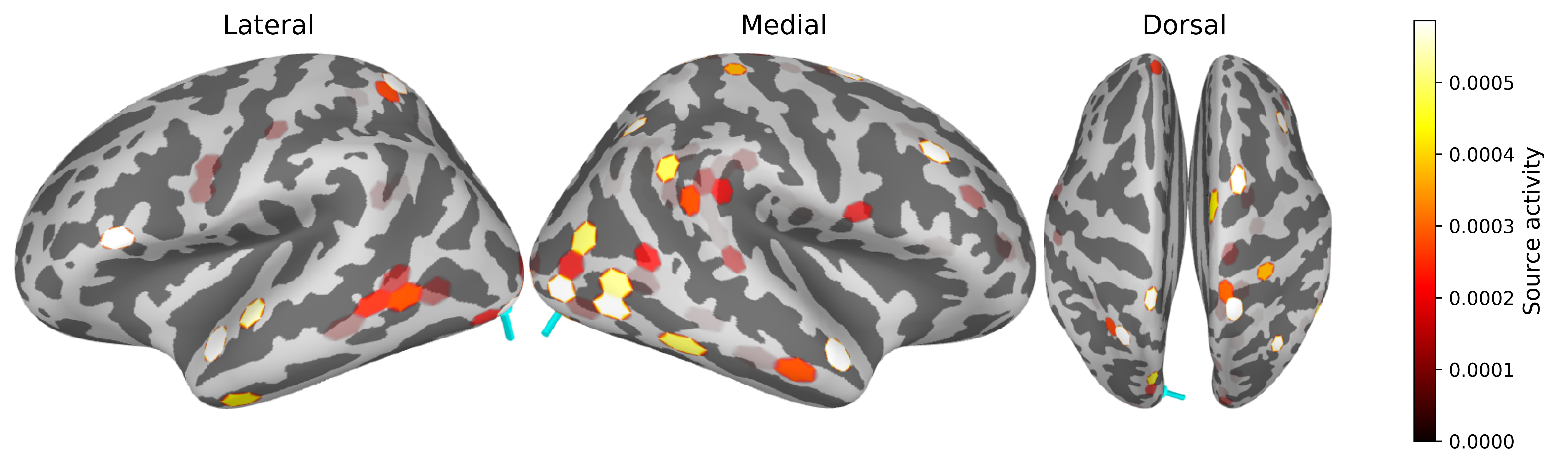}
        \caption{Convex Bounding (Single)}
    \end{subfigure}
    \hfill
    \begin{subfigure}[b]{0.48\linewidth}
        \centering
        \includegraphics[width=\linewidth]{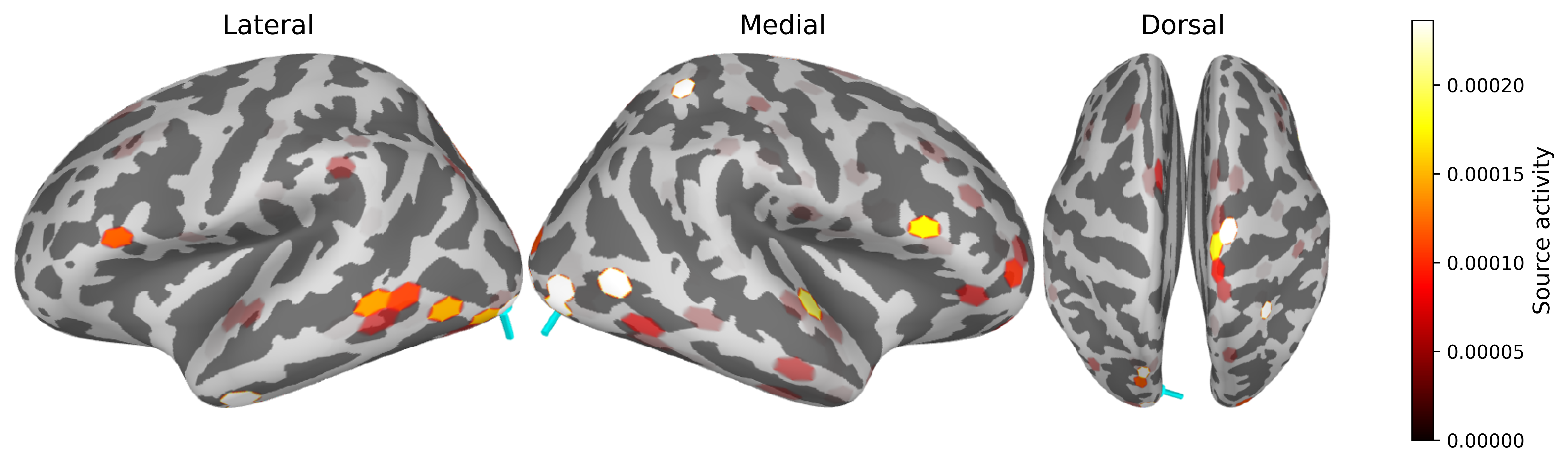}
        \caption{Convex Bounding (Avg)}
    \end{subfigure}

    \vspace{0.5em}

    \begin{subfigure}[b]{0.48\linewidth}
        \centering
        \includegraphics[width=\linewidth]{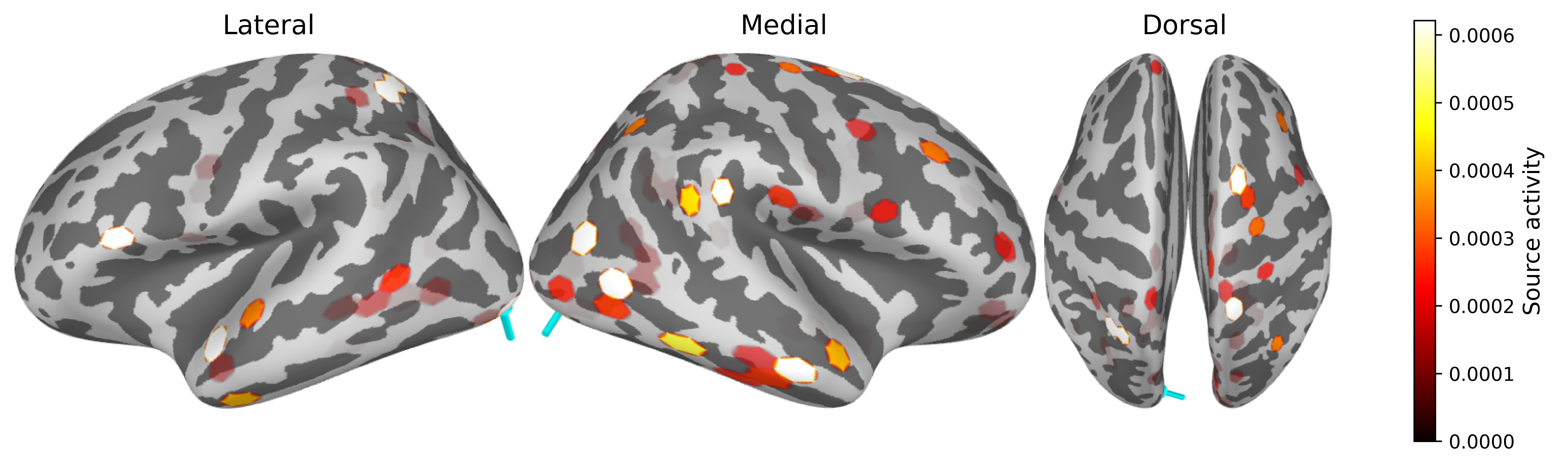}
        \caption{Deep CB (Single)}
    \end{subfigure}
    \hfill
    \begin{subfigure}[b]{0.48\linewidth}
        \centering
        \includegraphics[width=\linewidth]{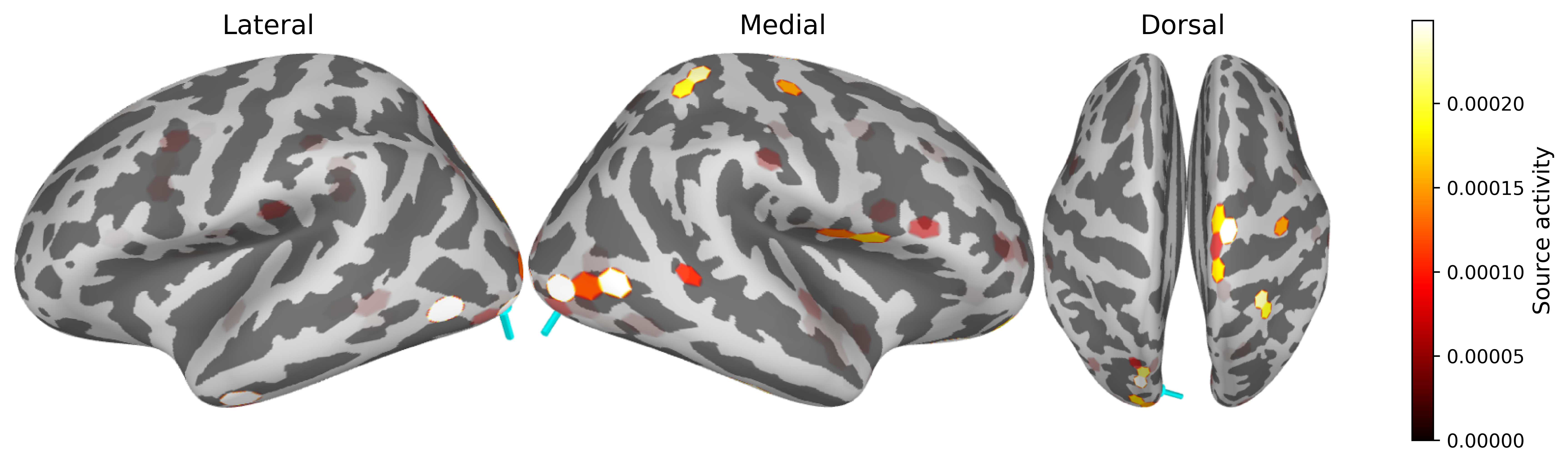}
        \caption{Deep CB (Avg)}
    \end{subfigure}

    \vspace{0.5em}

    \begin{subfigure}[b]{0.48\linewidth}
        \centering
        \includegraphics[width=\linewidth]{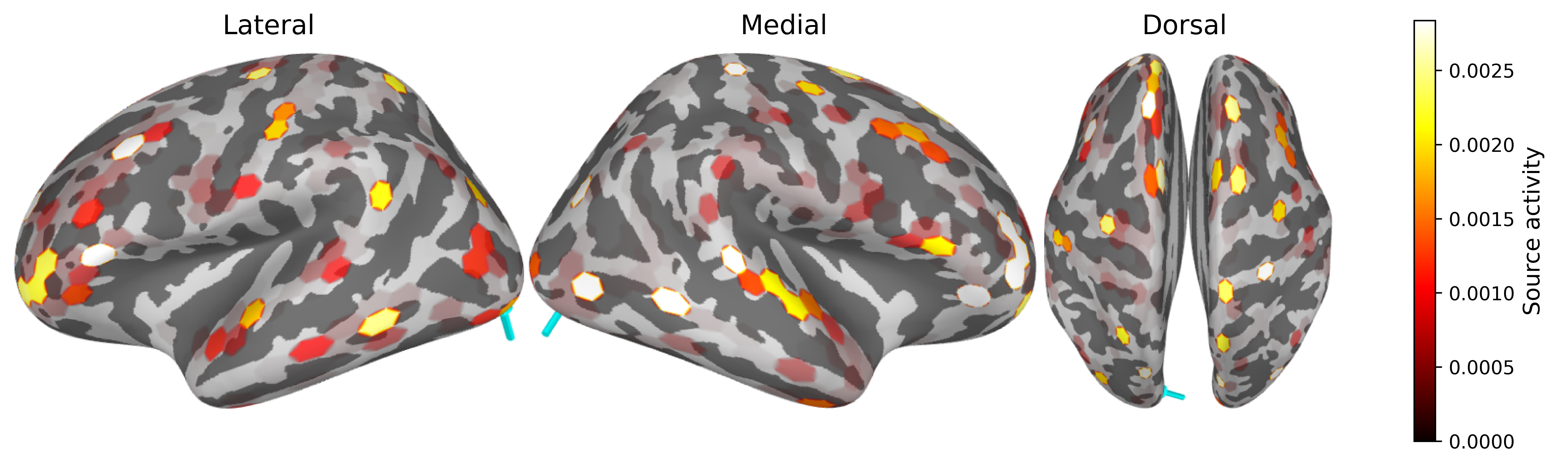}
        \caption{Deep Attn. CB (Single)}
    \end{subfigure}
    \hfill
    \begin{subfigure}[b]{0.48\linewidth}
        \centering
        \includegraphics[width=\linewidth]{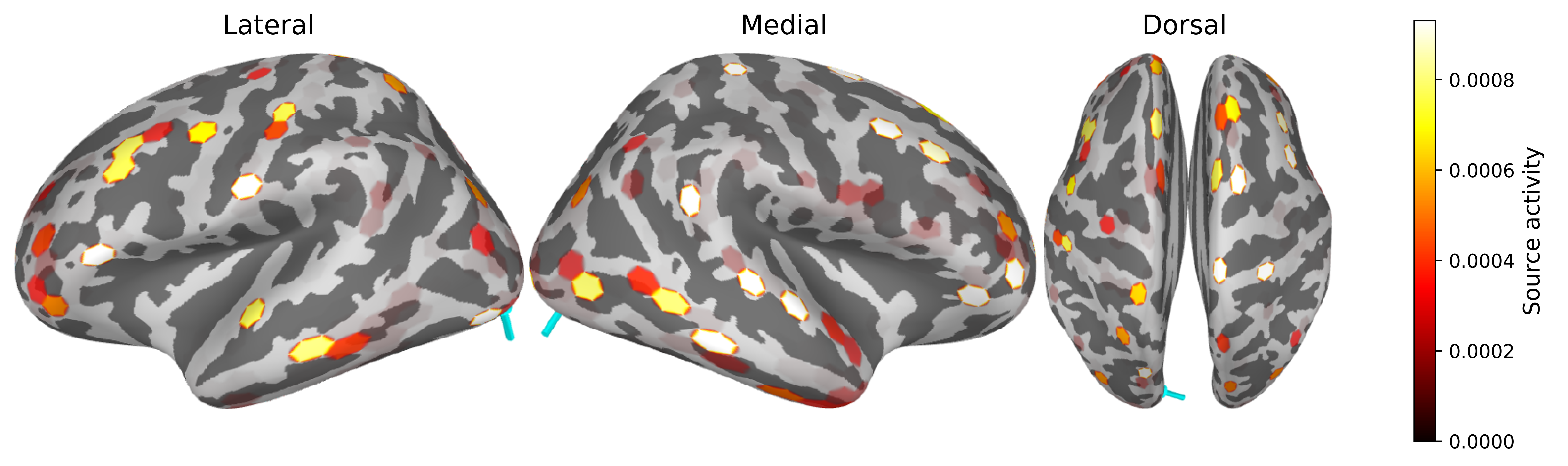}
        \caption{Deep Attn. CB (Avg)}
    \end{subfigure}

    \vspace{1em}

    \begin{subfigure}[b]{\linewidth}
        \centering
        \includegraphics[width=0.85\linewidth]{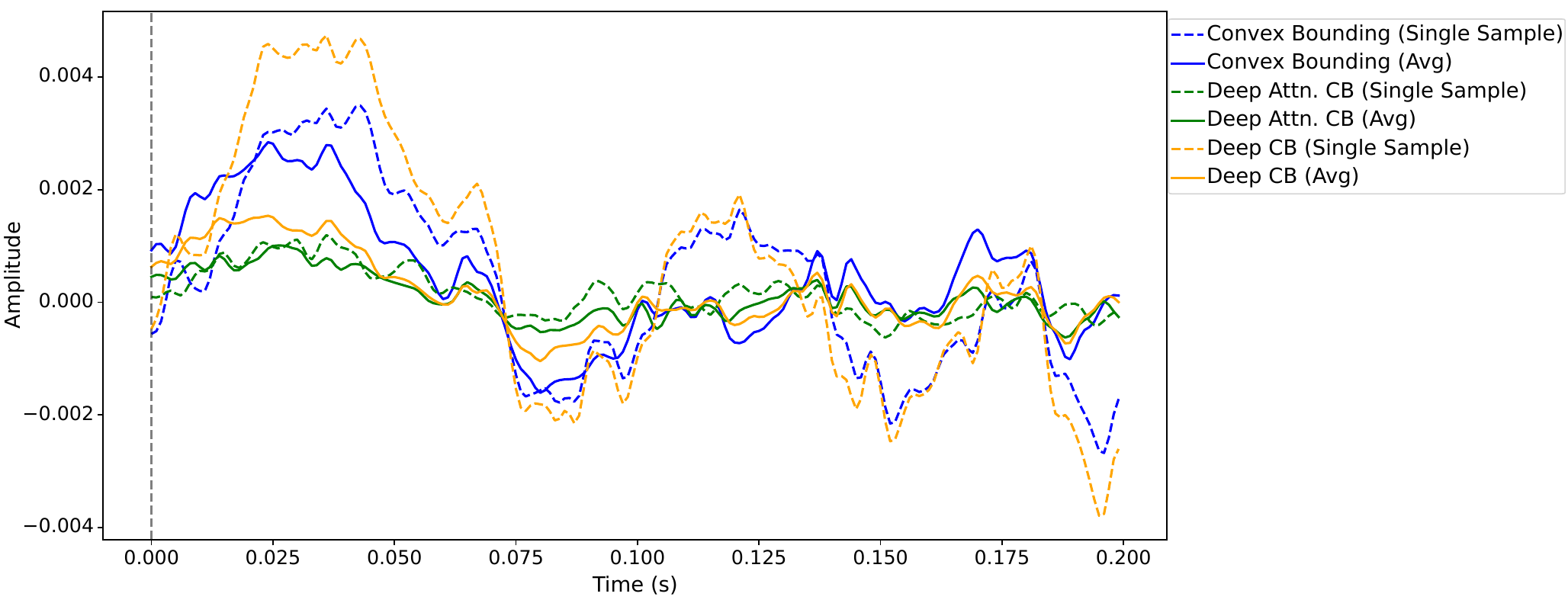}
        \caption{Temporal Dynamics of Peak Occipital Sources}
    \end{subfigure}

    \caption{Zero-shot real-world validation on the THINGS-EEG2 dataset. \textbf{(a--f)} Source localizations for early visual evoked potentials comparing highly noisy single-trial estimates (left column) against trial-averaged estimates (right column). Our unrolled architectures (Deep CB and Deep Attn. CB) successfully recover biologically plausible focal activity in the visual cortex. \textbf{(g)} The extracted temporal dynamics capture the characteristic early visual processing waveform across both single and averaged trials for the visual source marked with the arrow.}
    \label{fig:things_real_world}
\end{figure}

As shown in Figure~\ref{fig:things_real_world}, both our unrolled architectures successfully localize the evoked activity in the visual cortex. Notably, while the highly parameterized Deep Attn. CB dominates in simulated environments; its attention mechanism proves slightly more sensitive to the severe, uncalibrated noise of single-trial empirical data. Consequently, the less complex Deep CB yields the sharpest zero-shot single-trial localizations and suppresses uncalibrated sensor noise far better than the classical baseline, resulting in biologically plausible source estimates. Furthermore, the extracted temporal dynamics of the peak occipital sources successfully recover the characteristic waveform morphology of early visual processing.

\section{\rebuttal{Head-model generalization}}
\label{app:head_model_generalization}

\begin{figure}[H]
    \centering
    \includegraphics[width=0.96\linewidth]{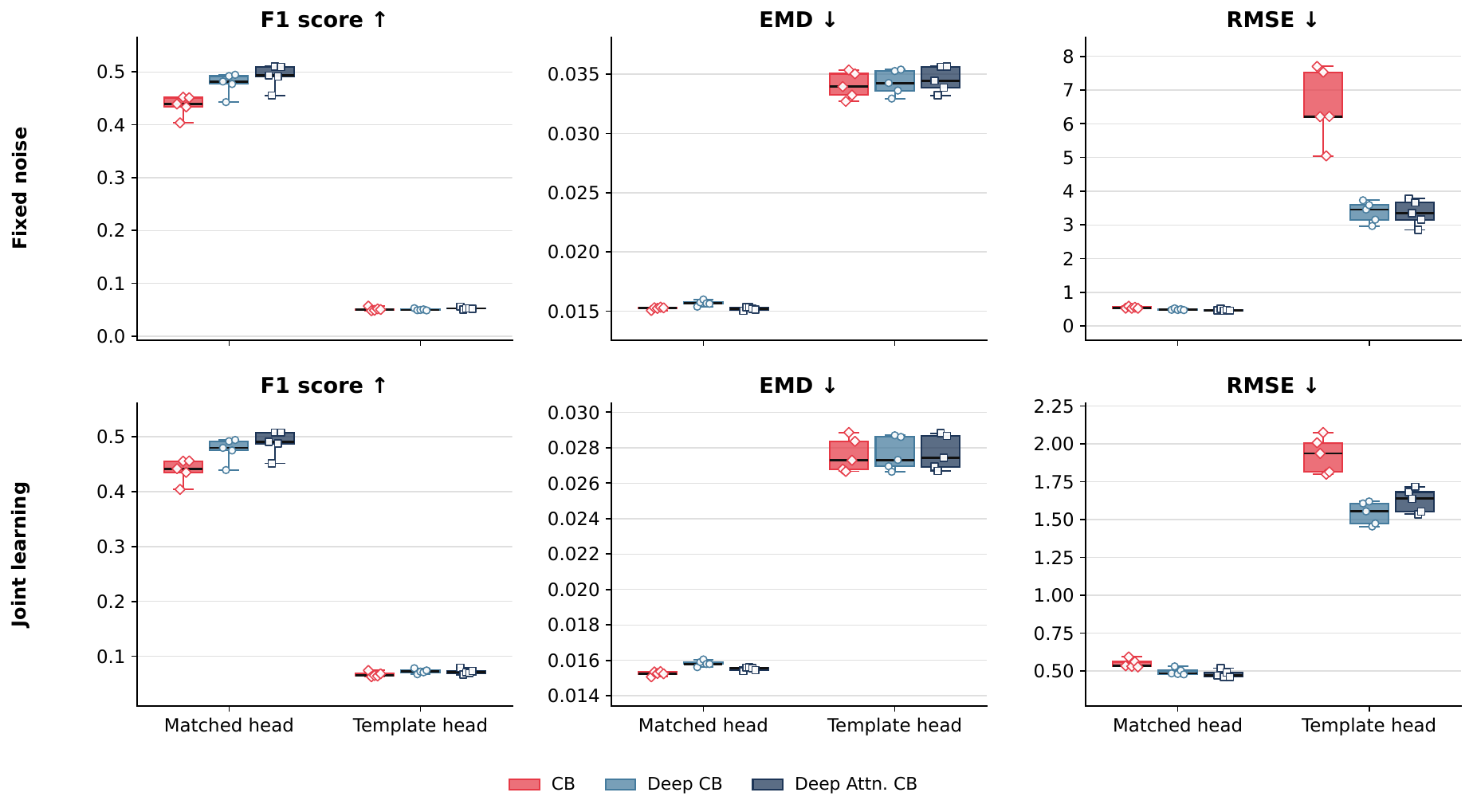}
    \caption{\rebuttal{Transfer to five unseen heads with matched subject-specific inference and with fsaverage-template inference. Points and error bars show subject-level mean $\pm$ SD. Template estimates were morphed to the individual source space before evaluation.}}
    \label{fig:head_model_generalization}
\end{figure}

\rebuttal{In this section, we test the transfer of our methods trained only on a single template head to MRI-derived, fixed-orientation BEM lead fields \cite{fuchs2001bem} from five previously unseen subjects \cite{nierhaus2021head}. Data-generation/inference conductivities for brain, skull, and scalp were $(0.332,0.0113,0.332)/(0.3,0.006,0.3)$~S/m. Results are subject-level mean $\pm$ SD, while we first aggregated five model seeds within each subject for learned methods.}

\rebuttal{The results in Figure \ref{fig:head_model_generalization} show a strong difference between matched and template anatomy. For the matched anatomy, the learned models consistently improve over their corresponding classical CB baselines, in terms of F1 and rMSE. For the harder no-individual-MRI condition, data were generated with each subject's validation lead field, inference used an fsaverage BEM with the corresponding sensor layout, and estimates were mapped into individual source space using an MNE surface morph. Template mismatch degraded every method, while the joint models can partly compensate for the mismatch in the forward model.
 Overall, the learned correction transfers to unseen individual heads when the inference anatomy is matched, but sparse prediction collapses when subject-specific leadfields are unavailable.}

\section{\rebuttal{THINGS-EEG2 cross-session consistency}}
\label{app:real_data_consistency}
\begin{figure}[H]
    \centering
    \includegraphics[width=\linewidth]{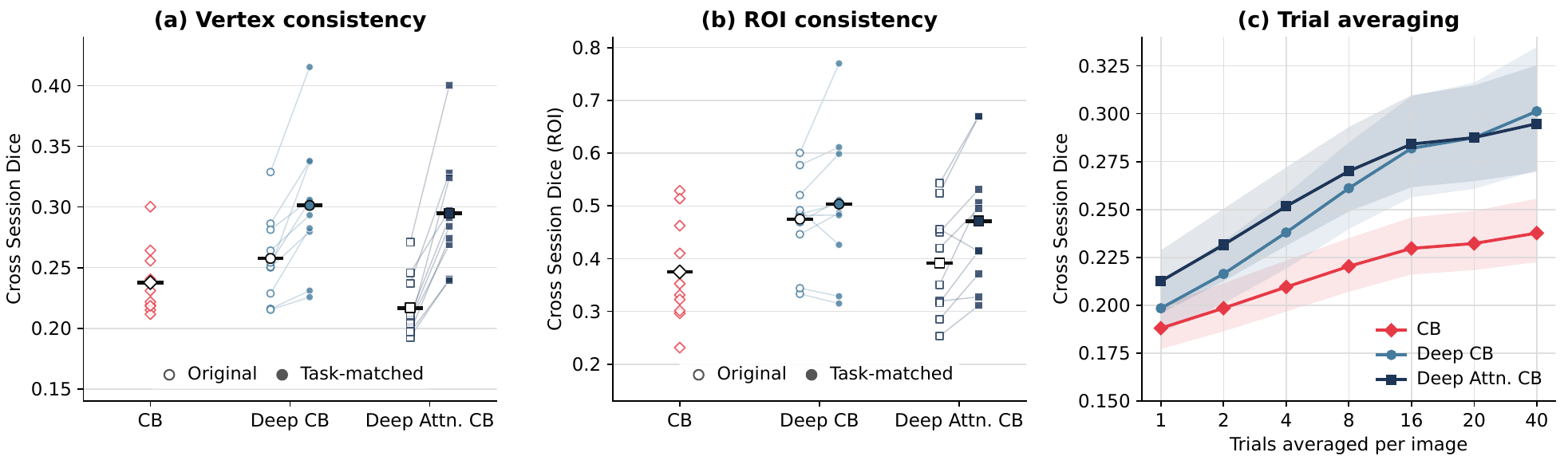}
    \caption{\rebuttal{THINGS-EEG2 cross-session consistency. Left and center: subject-level Top-5\% vertex and Top-4 ROI Dice for the frozen original models and models retrained only on task-matched simulations, not real-data fine-tuning. Right: Top-5\% vertex Dice versus the number of candidate trials averaged, with the independent reference fixed at 40 trials.}}
    \label{fig:things_consistency}
\end{figure}

\rebuttal{In Figure \ref{fig:things_consistency}, we quantified empirical repeatability on THINGS-EEG2 using 10 subjects, 200 repeated-image conditions, and 20 trials per image in each of four sessions \cite{gifford2022large}. Sessions 1-2 provided 40 candidate trials and sessions 3-4 an independent 40-trial reference. Trials were averaged in sensor space before applying each inverse once to improve the SNR. Each reconstruction was summarized by its vertex-wise RMS over 0-200~ms. We measured Dice overlap for the strongest 5\% of vertices (65 of 1,284) and, after averaging within 68 bilateral aparc regions, for the four strongest ROIs. Scores were averaged over images and then subjects, while uncertainty is between-subject SD.}

\rebuttal{While the frozen original models already outperform classical CB in both Vertex and ROI consistency, retraining the models on matched simulations using the THINGS fsaverage cap, an SNR range of $-10$ to $30$~dB, 200-ms signals, and 1,000-Hz sampling further improves source consistency. This demonstrates the value of matching simulations to the deployment setting.}

\rebuttal{When comparing trial averaging, for $n\in\{1,2,4,8,16,20,40\}$ candidate trials, the reference always used all 40 independent trials. For $n<40$, 50 shared random draws were averaged within each subject. At the equal four-trial budget, exact two-sided paired sign-flip tests with Holm correction showed improvements over CB for Deep CB ($+0.028$, unadjusted paired subject-bootstrap 95\% CI $[0.012,0.044]$, $p_{\mathrm{Holm}}=0.0117$) and Deep Attn. CB ($+0.042$, $[0.025,0.058]$, $p_{\mathrm{Holm}}=0.0078$). The results demonstrate consistent source detection across sessions, not localization accuracy, as ground-truth activations are unavailable.}

\section{\rebuttal{Transfer to MacKay Type-II updates}}
\label{app:mackay_transfer}
\begin{figure}[H]
    \centering
    \includegraphics[width=\linewidth]{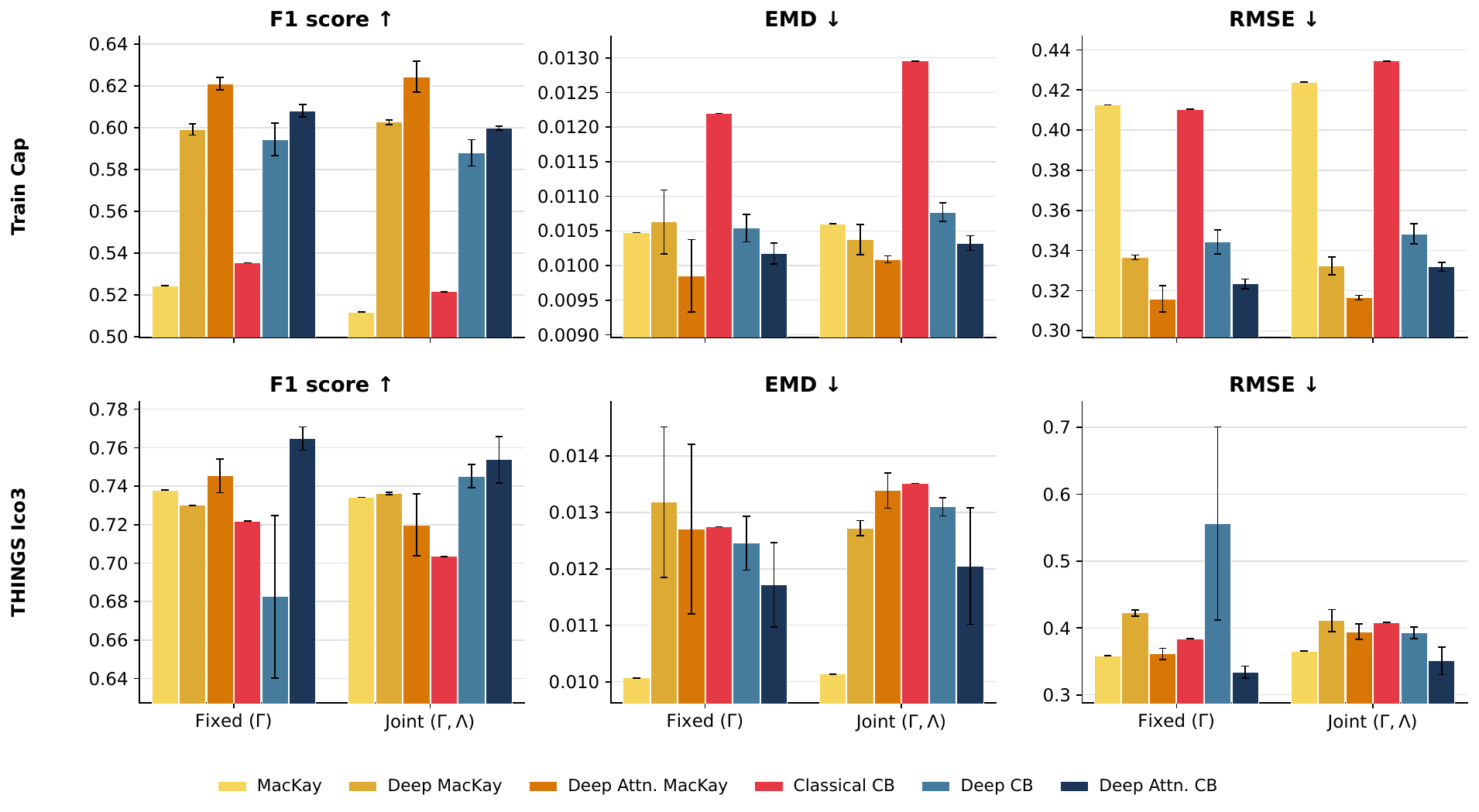}
    \caption{\rebuttal{Classical and learned MacKay updates on Train Cap and the shifted Things Ico3 setting. rMSE and F1 are shown separately for source-only and joint source/noise variants; both learned seeds are visible. The in-distribution gain is strong, whereas transfer under a sensor-layout shift is nonuniform.}}
    \label{fig:mackay_transfer}
\end{figure}
\rebuttal{To test whether correction learning is specific to convex bounding, we replaced the analytical source and noise steps by MacKay updates, $\gamma_n^{+}=e_n/(\gamma_n z_n)$ and $\lambda_m^{+}=d_m/(\lambda_m q_m)$, and learned the same pointwise or attention residual corrections around them. Classical MacKay was run for 250 iterations. Learned MacKay entries use 25 unfolded layers and two training seeds.}

\rebuttal{In Figure \ref{fig:mackay_transfer}, we compare the performance of the learned variants for both fixed noise and joint learning on both the Train Cap and THINGS generalization. On the Train Cap, we observe a large benefit of residual learning in comparison to classical MacKay updates, achieving results similar to Deep CB. For the unseen THINGS Ico3 setting, Deep Attn. MacKay performed slightly worse than Deep Attn. CB, while classical MacKay performed better in this setting. Notably, Deep CB showed less consistent behavior with large standard deviation across seeds. Overall, the results suggest that residual correction learning is not specific to Convex Bounding but can also improve other Type-II rules, especially in distribution.}

\section{\rebuttal{Zero-initialization diagnostic}}
\label{app:zero_initialization}
\begin{figure}[H]
    \centering
    \vspace{-0.1cm}
    \includegraphics[width=0.92\linewidth]{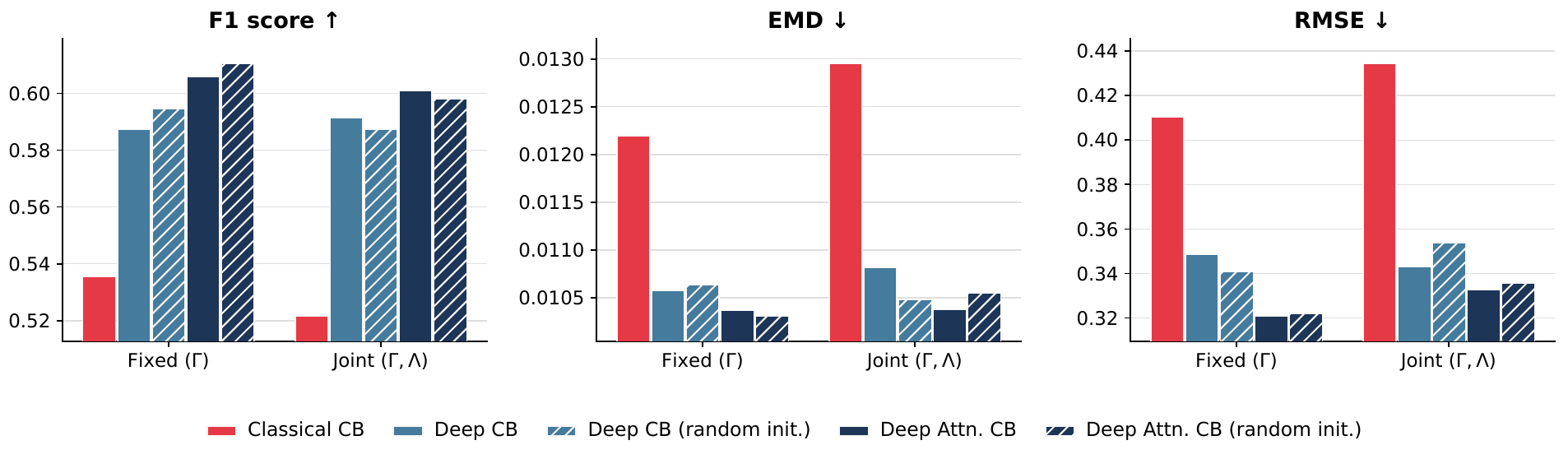}
    \caption{\rebuttal{Comparison of Deep CB and Deep Attn. CB trained with zero or random initialization for the residual correction.}}
    \label{fig:zero_initialization}
    \vspace{-0.3cm}
\end{figure}

\rebuttal{In Figure \ref{fig:zero_initialization}, we compare the proposed zero-output initialization with PyTorch's standard random initialization on the Train-Cap validation set. Both approaches achieve comparable performance, indicating that the model also trains successfully with standard initialization. We therefore use zero initialization for its structural interpretation rather than out of necessity for optimization.}

\section{\rebuttal{Type-II objective across layers}}
\label{app:typeii_objective}

\begin{figure}[H]
    \centering
    \vspace{-0.1cm}
    \includegraphics[width=\linewidth]{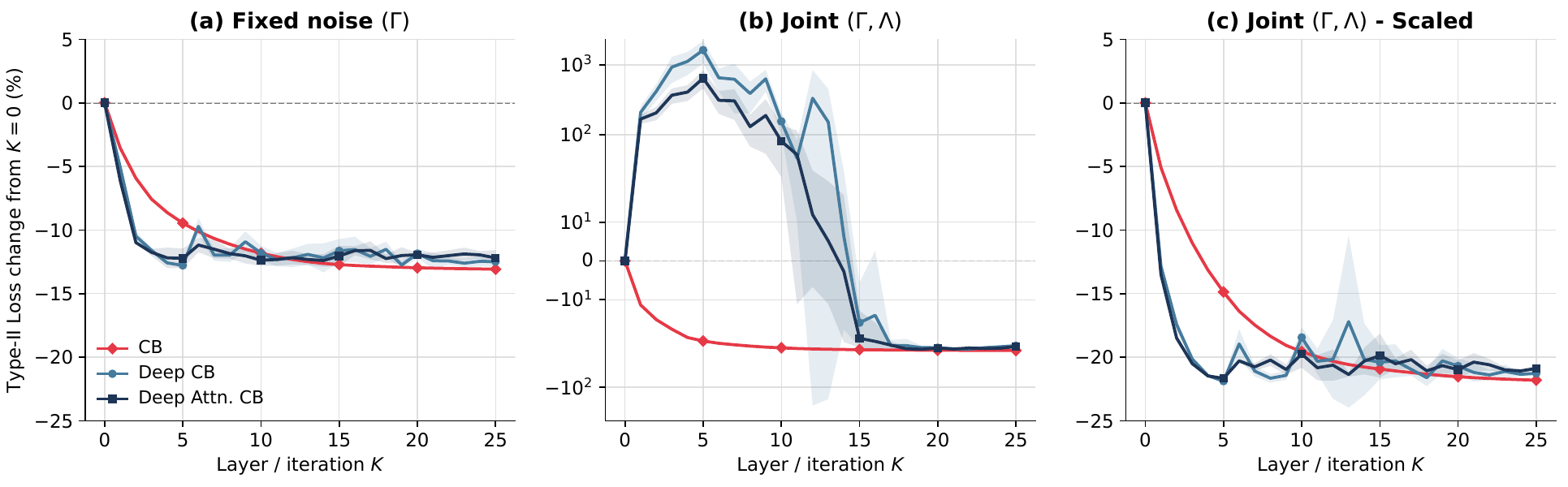}
    \caption{\rebuttal{Panel (a) shows that the learned CB variants rapidly decrease the Type-II loss over the first five layers, while afterwards exhibiting non-monotone updates. When $\boldsymbol{\Gamma}$ and $\boldsymbol{\Lambda}$ are learned jointly, the reconstruction loss permits a common rescaling of both quantities, explaining the large objective increases observed in panel (b). Applying the optimal post-hoc rescaling in panel (c) reveals that the joint variant likewise achieves a rapid decrease during the early iterations.}}
    \label{fig:typeii_objective}
    \vspace{-0.3cm}
\end{figure}

\rebuttal{In Figure~\ref{fig:typeii_objective}, we evaluate the Type-II objective in equation \ref{eq:typeII_loss} after each layer on 100 corrected Train-Cap validation samples. For the learned methods, results are aggregated across five model seeds. Classical CB exhibits a monotonic decrease by construction. The learned source-only models remain below their initial objective values at all reported layers, although the objective increases again after an initial rapid reduction. In contrast, the jointly learned models exhibit large early increases in the objective. This behavior arises because the source and noise covariances share a common scale degree of freedom. Let $\mathbf{C}=\mathbf{L}\boldsymbol{\Gamma} \mathbf{L}^\top+\boldsymbol{\Lambda}$ and $q=T^{-1}\sum_t \mathbf{y}(t)^\top \mathbf{C}^{-1}y(t)$. Under the common rescaling $(\boldsymbol{\Gamma},\boldsymbol{\Lambda})\mapsto(a\boldsymbol{\Gamma},a\boldsymbol{\Lambda})$, the objective becomes $M\log a+\log|\mathbf{C}|+q/a$ and is minimized at $a^\star=q/M$. Importantly, the posterior mean is invariant to this rescaling, since $(a\boldsymbol{\Gamma})\mathbf{L}^\top(a\mathbf{C})^{-1}\mathbf{Y}=\boldsymbol{\Gamma} \mathbf{L}^\top \mathbf{C}^{-1}\mathbf{Y}$, explaining the scaling of the neural network output. Applying this optimal rescaling as a post-hoc diagnostic in panel (c) repairs the spikes and leading also to rapid decrease in early iterations.}

\end{document}

%% file: math_commands.tex
\usepackage{amsmath,amsfonts,bm}
\usepackage{xspace}

\def\eqref#1{equation~\ref{#1}}

\def\1{\bm{1}}

\DeclareMathAlphabet{\mathsfit}{\encodingdefault}{\sfdefault}{m}{sl}
\SetMathAlphabet{\mathsfit}{bold}{\encodingdefault}{\sfdefault}{bx}{n}



%% file: references_R1.bib
@inproceedings{wipf2007analysis,
  author    = {Ram{\'i}rez, Rey and Palmer, Jason and Makeig, Scott and Rao, Bhaskar D. and Wipf, David P.},
  title     = {Analysis of Empirical Bayesian Methods for Neuroelectromagnetic Source Localization},
  booktitle = {Advances in Neural Information Processing Systems 19},
  pages     = {1505--1512},
  year      = {2007},
  publisher = {MIT Press},
  url       = {https://papers.nips.cc/paper_files/paper/2006/hash/ccd2e3eaa5c991ac880991328c8f1463-Abstract.html}
}

@article{wipf2009unified,
  author  = {Wipf, David and Nagarajan, Srikantan},
  title   = {A Unified Bayesian Framework for {MEG}/{EEG} Source Imaging},
  journal = {NeuroImage},
  volume  = {44},
  number  = {3},
  pages   = {947--966},
  year    = {2009},
  doi     = {10.1016/j.neuroimage.2008.02.059}
}

@article{wipf2010iterative,
  author  = {Wipf, David P. and Owen, J. P. and Attias, Hagai T. and Sekihara, Kensuke and Nagarajan, Srikantan S.},
  title   = {Robust Bayesian Estimation of the Location, Orientation, and Time Course of Multiple Correlated Neural Sources Using {MEG}},
  journal = {NeuroImage},
  volume  = {49},
  number  = {1},
  pages   = {641--655},
  year    = {2010},
  doi     = {10.1016/j.neuroimage.2009.06.083}
}

@article{cai2021robust,
  author  = {Cai, Chang and Hashemi, Ali and Diwakar, Mithun and Haufe, Stefan and Sekihara, Kensuke and Nagarajan, Srikantan S.},
  title   = {Robust Estimation of Noise for Electromagnetic Brain Imaging with the {Champagne} Algorithm},
  journal = {NeuroImage},
  volume  = {225},
  pages   = {117411},
  year    = {2021},
  doi     = {10.1016/j.neuroimage.2020.117411}
}

@article{tipping2001sparse,
  author  = {Tipping, Michael E.},
  title   = {Sparse Bayesian Learning and the Relevance Vector Machine},
  journal = {Journal of Machine Learning Research},
  volume  = {1},
  pages   = {211--244},
  year    = {2001}
}

@article{owen2012performance,
  author  = {Owen, J. P. and Wipf, David P. and Attias, Hagai T. and Sekihara, Kensuke and Nagarajan, Srikantan S.},
  title   = {Performance Evaluation of the {Champagne} Source Reconstruction Algorithm on Simulated and Real {M/EEG} Data},
  journal = {NeuroImage},
  volume  = {60},
  number  = {1},
  pages   = {305--323},
  year    = {2012},
  doi     = {10.1016/j.neuroimage.2011.12.027}
}

@article{hashemi2021unification,
  author  = {Hashemi, Ali and Cai, Chang and Kutyniok, Gitta and M{\"u}ller, Klaus-Robert and Nagarajan, Srikantan S. and Haufe, Stefan},
  title   = {Unification of Sparse Bayesian Learning Algorithms for Electromagnetic Brain Imaging with the Majorization Minimization Framework},
  journal = {NeuroImage},
  volume  = {239},
  pages   = {118309},
  year    = {2021},
  doi     = {10.1016/j.neuroimage.2021.118309}
}

@inproceedings{gregor2010learning,
  author    = {Gregor, Karol and LeCun, Yann},
  title     = {Learning Fast Approximations of Sparse Coding},
  booktitle = {Proceedings of the 27th International Conference on Machine Learning},
  pages     = {399--406},
  year      = {2010}
}

@article{hershey2014deep,
  author  = {Hershey, John R. and Le Roux, Jonathan and Weninger, Felix},
  title   = {Deep Unfolding: Model-Based Inspiration of Novel Deep Architectures},
  journal = {arXiv preprint arXiv:1409.2574},
  year    = {2014},
  doi     = {10.48550/arXiv.1409.2574}
}

@article{monga2021algorithm,
  author  = {Monga, Vishal and Li, Yuelong and Eldar, Yonina C.},
  title   = {Algorithm Unrolling: Interpretable, Efficient Deep Learning for Signal and Image Processing},
  journal = {IEEE Signal Processing Magazine},
  volume  = {38},
  number  = {2},
  pages   = {18--44},
  year    = {2021}
}

@article{peter2019learnedsbl,
  author  = {Peter, Rubin Jose and Murthy, Chandra R.},
  title   = {{Learned-SBL}: A Deep Learning Architecture for Sparse Signal Recovery},
  journal = {arXiv preprint arXiv:1909.08185},
  year    = {2019},
  doi     = {10.48550/arXiv.1909.08185}
}

@article{hecker2021convdip,
  author  = {Hecker, Lukas and Rupprecht, Rebekka and Tebartz van Elst, Ludger and Kornmeier, J{\"u}rgen},
  title   = {{ConvDip}: A Convolutional Neural Network for Better {EEG} Source Imaging},
  journal = {Frontiers in Neuroscience},
  volume  = {15},
  pages   = {569918},
  year    = {2021},
  doi     = {10.3389/fnins.2021.569918}
}

@article{pantazis2021deepmeg,
  author  = {Pantazis, Dimitrios and Adler, Amir},
  title   = {{MEG} Source Localization via Deep Learning},
  journal = {Sensors},
  volume  = {21},
  number  = {13},
  pages   = {4278},
  year    = {2021},
  doi     = {10.3390/s21134278}
}

@article{sun2022deepsif,
  author  = {Sun, Rui and Sohrabpour, Abbas and Worrell, Gregory A. and He, Bin},
  title   = {Deep Neural Networks Constrained by Neural Mass Models Improve Electrophysiological Source Imaging of Spatiotemporal Brain Dynamics},
  journal = {Proceedings of the National Academy of Sciences},
  volume  = {119},
  number  = {31},
  pages   = {e2201128119},
  year    = {2022},
  doi     = {10.1073/pnas.2201128119}
}

@article{liang2023sbl_dnn,
  author  = {Liang, Jiawen and Yu, Zhu Liang and Gu, Zhenghui and Li, Yuanqing},
  title   = {Electromagnetic Source Imaging With a Combination of Sparse Bayesian Learning and Deep Neural Network},
  journal = {IEEE Transactions on Neural Systems and Rehabilitation Engineering},
  volume  = {31},
  pages   = {2338--2348},
  year    = {2023},
  doi     = {10.1109/TNSRE.2023.3251420}
}

@article{morik2024physics3d,
  author={Morik, Marco and Hashemi, Ali and Müller, Klaus-Robert and Haufe, Stefan and Nakajima, Shinichi},
  journal={IEEE Transactions on Medical Imaging}, 
  title={Enhancing Brain Source Reconstruction by Initializing 3D Neural Networks With Physical Inverse Solutions}, 
  year={2026},
  volume={45},
  number={1},
  pages={231-242},
  doi={10.1109/TMI.2025.3594724}
}

@article{gifford2022large,
  author  = {Gifford, Alessandro T. and Dwivedi, Kshitij and Roig, Gemma and Cichy, Radoslaw M.},
  title   = {A Large and Rich {EEG} Dataset for Modeling Human Visual Object Recognition},
  journal = {NeuroImage},
  volume  = {264},
  pages   = {119754},
  year    = {2022}
}

@article{fuchs2002standardized,
  author  = {Fuchs, Manfred and Kastner, J{\"o}rn and Wagner, Michael and Hawes, Susan and Ebersole, John S.},
  title   = {A Standardized Boundary Element Method Volume Conductor Model},
  journal = {Clinical Neurophysiology},
  volume  = {113},
  number  = {5},
  pages   = {702--712},
  year    = {2002}
}

@article{GramfortEtAl2013a,
  author  = {Gramfort, Alexandre and Luessi, Martin and Larson, Eric and Engemann, Denis A. and Strohmeier, Daniel and Brodbeck, Christian and Goj, Roman and Jas, Mainak and Brooks, Teon and Parkkonen, Lauri and H{\"a}m{\"a}l{\"a}inen, Matti S.},
  title   = {{MEG} and {EEG} Data Analysis with {MNE}-{Python}},
  journal = {Frontiers in Neuroscience},
  volume  = {7},
  number  = {267},
  pages   = {1--13},
  year    = {2013}
}

@article{fischl1999high,
  author  = {Fischl, Bruce and Sereno, Martin I. and Tootell, Roger B. H. and Dale, Anders M.},
  title   = {High-Resolution Intersubject Averaging and a Coordinate System for the Cortical Surface},
  journal = {Human Brain Mapping},
  volume  = {8},
  number  = {4},
  pages   = {272--284},
  year    = {1999}
}

@book{colton1998inverse,
  author    = {Colton, David L. and Kress, Rainer},
  title     = {Inverse Acoustic and Electromagnetic Scattering Theory},
  volume    = {93},
  publisher = {Springer},
  year      = {1998}
}

@article{kaipio2007statistical,
  author  = {Kaipio, Jari and Somersalo, Erkki},
  title   = {Statistical Inverse Problems: Discretization, Model Reduction and Inverse Crimes},
  journal = {Journal of Computational and Applied Mathematics},
  volume  = {198},
  number  = {2},
  pages   = {493--504},
  year    = {2007}
}

@article{wolters2010combined,
  author  = {Wolters, Carsten H. and Lew, Seok and MacLeod, Rob S. and H{\"a}m{\"a}l{\"a}inen, Matti},
  title   = {Combined {EEG}/{MEG} Source Analysis Using Calibrated Finite Element Head Models},
  journal = {Biomedizinische Technik/Biomedical Engineering},
  volume  = {55},
  number  = {Suppl 1},
  pages   = {64--68},
  year    = {2010}
}

@article{pascual2002standardized,
  author  = {Pascual-Marqui, Roberto D.},
  title   = {Standardized Low-Resolution Brain Electromagnetic Tomography ({sLORETA}): Technical Details},
  journal = {Methods and Findings in Experimental and Clinical Pharmacology},
  volume  = {24},
  number  = {Suppl D},
  pages   = {5--12},
  year    = {2002}
}

@book{sekihara2015bayesian,
  author    = {Sekihara, Kensuke and Nagarajan, Srikantan S.},
  title     = {Electromagnetic Brain Imaging: A Bayesian Perspective},
  publisher = {Springer International Publishing},
  address   = {Cham},
  year      = {2015},
  doi       = {10.1007/978-3-319-14947-9}
}

@inproceedings{lee2015deeply,
  author    = {Lee, Chen-Yu and Xie, Saining and Gallagher, Patrick and Zhang, Zhengyou and Tu, Zhuowen},
  title     = {Deeply-Supervised Nets},
  booktitle = {Proceedings of the Eighteenth International Conference on Artificial Intelligence and Statistics},
  series    = {Proceedings of Machine Learning Research},
  volume    = {38},
  pages     = {562--570},
  year      = {2015},
  publisher = {PMLR},
  url       = {https://proceedings.mlr.press/v38/lee15a.html}
}

@inproceedings{vaswani2017attention,
  author    = {Vaswani, Ashish and Shazeer, Noam and Parmar, Niki and Uszkoreit, Jakob and Jones, Llion and Gomez, Aidan N. and Kaiser, {\L}ukasz and Polosukhin, Illia},
  title     = {Attention Is All You Need},
  booktitle = {Advances in Neural Information Processing Systems 30},
  year      = {2017},
  publisher = {Curran Associates, Inc.},
  url       = {https://papers.nips.cc/paper/7181-attention-is-all-you-need}
}

@article{hashemi2024fullstructure,
  title   = {Joint Learning of Full-Structure Noise in Hierarchical Bayesian Regression Models},
  author  = {Hashemi, Ali and Cai, Chang and Gao, Yijing and Ghosh, Sanjay and M{\"u}ller, Klaus-Robert and Nagarajan, Srikantan S. and Haufe, Stefan},
  journal = {IEEE Transactions on Medical Imaging},
  volume  = {43},
  number  = {2},
  pages   = {610--624},
  year    = {2024},
  doi     = {10.1109/TMI.2022.3224085}
}

@inproceedings{hashemi2021spatiotemporal,
  title     = {Efficient Hierarchical Bayesian Inference for Spatio-Temporal Regression Models in Neuroimaging},
  author    = {Hashemi, Ali and Gao, Yijing and Cai, Chang and Ghosh, Sanjay and M{\"u}ller, Klaus-Robert and Nagarajan, Srikantan S. and Haufe, Stefan},
  booktitle = {Advances in Neural Information Processing Systems},
  volume    = {34},
  pages     = {24855--24870},
  year      = {2021}
}

@inproceedings{wipf2007newview,
  title     = {A New View of Automatic Relevance Determination},
  author    = {Wipf, David P. and Nagarajan, Srikantan S.},
  booktitle = {Advances in Neural Information Processing Systems 20},
  pages     = {1625--1632},
  year      = {2007}
}

@article{wipf2004sparse,
  title   = {Sparse Bayesian Learning for Basis Selection},
  author  = {Wipf, David P. and Rao, Bhaskar D.},
  journal = {IEEE Transactions on Signal Processing},
  volume  = {52},
  number  = {8},
  pages   = {2153--2164},
  year    = {2004},
  doi     = {10.1109/TSP.2004.831016}
}

@article{ji2008bayesian,
  title   = {Bayesian Compressive Sensing},
  author  = {Ji, Shihao and Xue, Ya and Carin, Lawrence},
  journal = {IEEE Transactions on Signal Processing},
  volume  = {56},
  number  = {6},
  pages   = {2346--2356},
  year    = {2008},
  doi     = {10.1109/TSP.2007.914345}
}

@article{gerstoft2017sparse,
  title   = {Sparse Bayesian Learning for DOA Estimation in Heteroscedastic Noise},
  author  = {Gerstoft, Peter and Nannuru, Santosh and Mecklenbr{\"a}uker, Christoph F. and Leus, Geert},
  journal = {arXiv preprint arXiv:1711.03847},
  year    = {2017}
}

@article{megevand2020electric,
  author  = {M{\'e}gevand, Pierre and Seeck, Margitta},
  title   = {Electric Source Imaging for Presurgical Epilepsy Evaluation: Current Status and Future Prospects},
  journal = {Expert Review of Medical Devices},
  volume  = {17},
  number  = {5},
  pages   = {405--412},
  year    = {2020},
  doi     = {10.1080/17434440.2020.1748008}
}

@article{nierhaus2021head,
  author  = {Nierhaus, Till and Vidaurre, Carmen and Sannelli, Claudia and M{\"u}ller, Klaus-Robert and Villringer, Arno},
  title   = {Immediate Brain Plasticity after One Hour of Brain--Computer Interface ({BCI})},
  journal = {The Journal of Physiology},
  volume  = {599},
  number  = {9},
  pages   = {2435--2451},
  year    = {2021},
  doi     = {10.1113/JP278118}
}

@article{hu2022ddpg,
  author  = {Hu, Qiyu and Shi, Shuhan and Cai, Yunlong and Yu, Guanding},
  title   = {{DDPG}-Driven Deep-Unfolding With Adaptive Depth for Channel Estimation With Sparse Bayesian Learning},
  journal = {IEEE Transactions on Signal Processing},
  volume  = {70},
  pages   = {4665--4680},
  year    = {2022},
  doi     = {10.1109/TSP.2022.3207269}
}

@inproceedings{gao2023ampsbl,
  author  = {Gao, Jiabao and Zhong, Caijun and Li, Geoffrey Ye},
  title   = {{AMP-SBL} Unfolding for Wideband {mmWave} Massive {MIMO} Channel Estimation},
  booktitle={2023 IEEE International Conference on Communications Workshops (ICC Workshops)},
  pages={60--65},
  year={2023},
  doi={10.1109/ICCWorkshops57953.2023.10283596}
}

@article{gong2023sblnet,
  author  = {Gong, Zhenghui and Su, Xiaolong and Hu, Panhe and Liu, Shuowei and Liu, Zhen},
  title   = {Deep Unfolding Sparse Bayesian Learning Network for Off-Grid {DOA} Estimation with Nested Array},
  journal = {Remote Sensing},
  volume  = {15},
  number  = {22},
  pages   = {5320},
  year    = {2023},
  doi     = {10.3390/rs15225320}
}

@article{balaji2026deepunfolding,
  author={Balaji, Rushabha and Chen, Kuan-Lin and Cabric, Danijela and Rao, Bhaskar D.},
  journal={IEEE Transactions on Signal Processing},
  title={Sparse Bayesian Learning Algorithms Revisited: From Learning Majorizers to Structured Algorithmic Learning Using Neural Networks},
  year={2026},
  volume={74},
  pages={2476-2490},
  doi={10.1109/TSP.2026.3699618}
}

@article{fuchs2001bem,
  title   = {Boundary element method volume conductor models for EEG source reconstruction},
  author  = {Fuchs, Manfred and Wagner, Michael and Kastner, J{\"o}rn},
  journal = {Clinical Neurophysiology},
  volume  = {112},
  number  = {8},
  pages   = {1400--1407},
  year    = {2001},
  doi     = {10.1016/S1388-2457(01)00589-2}
}

@article{gramfort2012mixed,
  title={Mixed-norm estimates for the M/EEG inverse problem using accelerated gradient methods},
  author={Gramfort, Alexandre and Kowalski, Matthieu and H{\"a}m{\"a}l{\"a}inen, Matti},
  journal={Physics in medicine and biology},
  volume={57},
  number={7},
  pages={1937--1961},
  year={2012},
  publisher={IOP Publishing}
}

@inproceedings{he2016deep,
  title     = {Deep Residual Learning for Image Recognition},
  author    = {He, Kaiming and Zhang, Xiangyu and Ren, Shaoqing
               and Sun, Jian},
  booktitle = {Proceedings of the IEEE Conference on Computer
               Vision and Pattern Recognition},
  pages     = {770--778},
  year      = {2016}
}

@inproceedings{um2020solver,
  title     = {Solver-in-the-Loop: Learning from Differentiable
               Physics to Interact with Iterative {PDE}-Solvers},
  author    = {Um, Kiwon and Brand, Robert and Fei, Yun (Raymond)
               and Holl, Philipp and Thuerey, Nils},
  booktitle = {Advances in Neural Information Processing Systems},
  volume    = {33},
  year      = {2020}
}
